\documentclass{fairmeta}

\usepackage{multirow}
\usepackage{tabularx, booktabs} %% Load packages that you use
\usepackage{amsmath}
\usepackage[normalem]{ulem}
\usepackage{xspace}
\usepackage{pifont}% http://ctan.org/pkg/pifont

\def\bf{\bfseries\sffamily}

\newcommand{\kindatiny}{\fontsize{6pt}{7.2pt}\selectfont}

\newlength\savewidth

\newcommand{\tablestyle}[2]{%
    \fontfamily{ptm}\selectfont%
    \let\itold\it%
    \def\it{\itold \fontfamily{ptm}\selectfont}%
    \setlength{\tabcolsep}{#1}\renewcommand{\arraystretch}{#2}\centering\kindatiny%
    \let\citeold\cite%
    \renewcommand{\cite}[1]{\normalfont\fontfamily{ptm}\selectfont\tiny\citeold{##1}}%
}
\newcommand{\bigtablestyle}[2]{%
    \fontfamily{ptm}\selectfont%
    \let\itold\it%
    \def\it{\itold \fontfamily{ptm}\selectfont}%
    \setlength{\tabcolsep}{#1}\renewcommand{\arraystretch}{#2}\centering\footnotesize%
    \let\citeold\cite%
    \renewcommand{\cite}[1]{\normalfont\fontfamily{ptm}\selectfont\footnotesize\citeold{##1}}%
}

\renewcommand{\paragraph}[1]{\vspace{1.25mm}\noindent\textbf{#1}}

\newcolumntype{x}[1]{>{\centering\arraybackslash}p{#1pt}}
\newcolumntype{y}[1]{>{\raggedright\arraybackslash}p{#1pt}}
\newcolumntype{z}[1]{>{\raggedleft\arraybackslash}p{#1pt}}
\newcolumntype{w}{>{\centering\arraybackslash}p{18pt}}
\newcolumntype{a}{>{\centering\arraybackslash}p{16pt}}

\definecolor{c0-title-bkg}{HTML}{ffffff}
\definecolor{c0-title-text}{HTML}{000000}
\definecolor{c0-item-bkg}{HTML}{ffffff}
\definecolor{c0-item-text}{HTML}{676767}

\definecolor{c1-title-bkg}{HTML}{d1e2dd}
\definecolor{c1-title-text}{HTML}{005953}
\definecolor{c1-item-bkg}{HTML}{e6efec}
\definecolor{c1-item-text}{HTML}{2d7b6d}

\definecolor{c2-title-bkg}{HTML}{cfe1e1}
\definecolor{c2-title-text}{HTML}{005760}
\definecolor{c2-item-bkg}{HTML}{e4eeed}
\definecolor{c2-item-text}{HTML}{24797b}

\definecolor{c3-title-bkg}{HTML}{cddfe5}
\definecolor{c3-title-text}{HTML}{00536b}
\definecolor{c3-item-bkg}{HTML}{e2ecef}
\definecolor{c3-item-text}{HTML}{287687}

\definecolor{c4-title-bkg}{HTML}{cedce8}
\definecolor{c4-title-text}{HTML}{124e74}
\definecolor{c4-item-bkg}{HTML}{e1eaf1}
\definecolor{c4-item-text}{HTML}{3a7190}

\definecolor{c5-title-bkg}{HTML}{d0d9eb}
\definecolor{c5-title-text}{HTML}{324779}
\definecolor{c5-item-bkg}{HTML}{e1e8f3}
\definecolor{c5-item-text}{HTML}{4d6b97}

\definecolor{c6-title-bkg}{HTML}{d3d5ed}
\definecolor{c6-title-text}{HTML}{493e7b}
\definecolor{c6-item-bkg}{HTML}{e3e5f5}
\definecolor{c6-item-text}{HTML}{61639b}

\definecolor{c7-title-bkg}{HTML}{dad1ed}
\definecolor{c7-title-text}{HTML}{5a3477}
\definecolor{c7-item-bkg}{HTML}{e5e1f5}
\definecolor{c7-item-text}{HTML}{725b99}

\definecolor{c8-title-bkg}{HTML}{ded1ec}
\definecolor{c8-title-text}{HTML}{633273}
\definecolor{c8-item-bkg}{HTML}{ebe2f6}
\definecolor{c8-item-text}{HTML}{7c5997}

\definecolor{c9-title-bkg}{HTML}{e5d1eb}
\definecolor{c9-title-text}{HTML}{6c2f6b}
\definecolor{c9-item-bkg}{HTML}{f0e0f6}
\definecolor{c9-item-text}{HTML}{885591}

\definecolor{c10-title-bkg}{HTML}{ebd1e7}
\definecolor{c10-title-text}{HTML}{722e5f}
\definecolor{c10-item-bkg}{HTML}{f5e2f3}
\definecolor{c10-item-text}{HTML}{915487}

\definecolor{avg-title-bkg}{HTML}{f3f3f3}
\definecolor{avg-title-text}{HTML}{000000}
\definecolor{avg-item-bkg}{HTML}{f3f3f3}
\definecolor{avg-item-text}{HTML}{000000}

\definecolor{promptcolor}{HTML}{D1D0F2}
\definecolor{promptcolorheader}{HTML}{bdbcec}

\ExplSyntaxOn

\cs_generate_variant:Nn \coffin_typeset:Nnnnn {Nffff}

\NewDocumentCommand\rotbox{ O{l,H} D<>{0pt,0pt} m m}{
    \hcoffin_set:Nn \l_tmpa_coffin {#4}
    \coffin_rotate:Nn \l_tmpa_coffin {#3}
    \coffin_typeset:Nffff \l_tmpa_coffin 
        {\clist_item:nn{#1}{1}}
        {\clist_item:nn{#1}{2}}
        {\clist_item:nn{#2}{1}}
        {\clist_item:nn{#2}{2}}
}
\ExplSyntaxOff

\newlength{\ccustomlen}
\newcommand{\ccustom}[3][c0]{%
    \cellcolor{#1-item-bkg}{%
        \rotbox[l,t]{90}{%
            \parbox[t]{\ccustomlen}{%
                \ifthenelse{\isempty{#3}}{%
                    \mbox{%
                        \kindatiny\textcolor{#1-title-text}{#2}%
                    }%
                }{%
                    \kindatiny\textcolor{#1-title-text}{#2} \\%
                    \tiny{\textcolor{#1-item-text}{\it #3}}%
                }%
            }%
        }%
    }%
}

\makeatother

\usepackage{amssymb}
\usepackage{xifthen}
\usepackage{wrapfig}
\usepackage{makecell}
\usepackage[percent]{overpic}
\usepackage{multicol}
\usepackage{hyperref}
\usepackage{bbding}
\usepackage{enumitem}

\title{OmniStream: Mastering Perception, Reconstruction and Action in Continuous Streams}

\newcommand{\model}{OmniStream\xspace}
\definecolor{skyblue}{RGB}{203, 221, 245}
\definecolor{highlightcolor}{RGB}{250, 235, 215}
\newcommand{\skyblue}{\rowcolor{skyblue}}

\author[1,2,*]{Yibin Yan}
\author[3,*]{Jilan Xu}
\author[1]{Shangzhe Di}
\author[1]{Haoning Wu}
\author[1]{Weidi Xie}

\affiliation[1]{School of Artificial Intelligence, SJTU}
\affiliation[2]{Shanghai Innovation Institute}
\affiliation[3]{VGG, Oxford}

\contribution[*]{Joint first author}

\abstract{
Modern visual agents require representations that are general, causal, and physically structured to operate in real-time streaming environments.
However, current vision foundation models remain fragmented, specializing narrowly in image semantic perception, offline temporal modeling, or spatial geometry. This paper introduces \textbf{\model}, a unified streaming visual backbone that effectively perceives, reconstructs, and acts from diverse visual inputs. By incorporating causal spatiotemporal attention and 3D rotary positional embeddings (3D-RoPE), our model supports efficient, frame-by-frame online processing of video streams via a persistent KV-cache.
We pre-train \textbf{\model} using a synergistic multi-task framework coupling static and temporal representation learning, streaming geometric reconstruction, and vision-language alignment on 29 datasets.
Extensive evaluations show that, even with a strictly frozen backbone, \textbf{\model} achieves consistently competitive performance with specialized experts across image and video probing, streaming geometric reconstruction, complex video and spatial reasoning, as well as robotic manipulation~(unseen at training). Rather than pursuing benchmark-specific dominance, our work demonstrates the viability of training a single, versatile vision backbone that generalizes across semantic, spatial, and temporal reasoning, {\em i.e.}, a more meaningful step toward general-purpose visual understanding for interactive and embodied agents.
}

\metadata[Page]{\url{https://go2heart.github.io/omnistream}}
\metadata[Code]{\url{https://github.com/Go2Heart/OmniStream}}

\begin{document}

\maketitle

\section{Introduction}
\begin{figure}[t]
    \centering
    \includegraphics[width=1.0\textwidth]{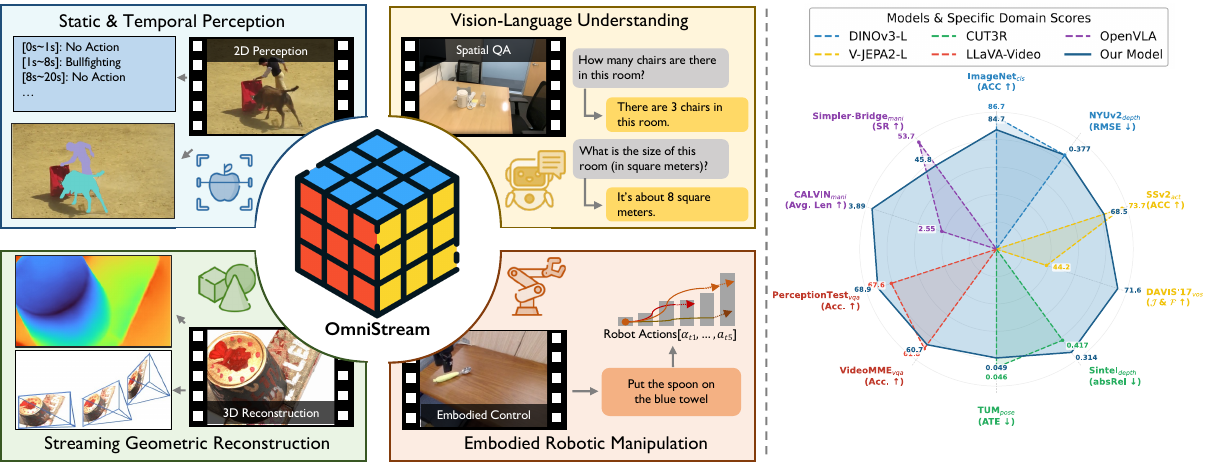}
    \caption{Left: \textbf{\model} supports a wide spectrum of tasks, including 2D/3D perception, vision-language understanding, and embodied robotic manipulation. 
    Right: the frozen features of our single backbone achieve highly competitive or superior performance compared to leading domain-specific experts.
    }
    \label{fig:teaser}
\end{figure}
Modern visual agents are increasingly deployed in \emph{streaming} settings: a robot observes the world through a camera, an assistant watches a video, or an AR device processes an egocentric feed. In such cases, the agent must update its beliefs online from a continuous stream, under tight latency and memory budgets.
A useful representation must therefore be (i) \emph{general}, supporting recognition, reasoning, and interaction; (ii) \emph{causal}, depending only on past and present frames; and (iii) \emph{structured}, capturing not only appearance but also geometry and motion.

In the language community, models achieve generality with a single autoregressive backbone trained under next-token prediction~\cite{gpt3,gpt4,super-instruction}.
This works not merely due to scale, but because the diverse tasks can share a common representation and training interface. 
Vision, however, remains far less unified. 
The tasks differ not only in supervision but also in the nature of their outputs: discrete labels~\cite{imagenet}, segmentation masks~\cite{sam}, dense depth~\cite{depthanything1}, 3D geometry~\cite{vggt}, and temporally evolving predictions. As a result, the field has converged on specialized foundation models: image encoders ({\em e.g.}, DINO~\cite{dinov2,dinov3}, SigLIP~\cite{siglip,siglip2}), video models ({\em e.g.}, V-JEPA~\cite{vjepa,vjepa2}, VideoMAE~\cite{videomae,videomae2}), and geometric experts ({\em e.g.}, DepthAnything~\cite{depthanything1,depthanything2,depthanything3}, VGGT~\cite{vggt}). While effective in-domain, these models typically learn representations tailored to a narrow objective (semantic invariances, temporal motion, or spatial geometry), and do not directly yield a single backbone that transfers across \emph{static semantics}, \emph{temporal dynamics}, and \emph{3D structure}---especially in an online, causal regime.

A natural direction is to ``unify'' vision by converting everything into text-like token generation, as explored by Florence~\cite{florence,florence2}, OFA~\cite{ofa}, and Unified-IO~\cite{unified-io,unified-io2}. These models provide a convenient interface, but the unification still largely happens at the \emph{output} level. In practice, accommodating new tasks often requires expensive re-training, re-tokenization of outputs, or architectural adjustments to the generative head. This leaves an open question that is more representation-centric than interface-centric: Can we learn a \emph{single streaming visual backbone} whose representations are sufficiently universal that many downstream tasks can be solved on top of it, without modifying or fine-tuning the backbone?

In this paper, we propose \textbf{\model}, a unified streaming visual backbone that turns a strong pre-trained image model into an online, temporally causal model while preserving its spatial priors. \model is built around two design choices. First, we introduce \textbf{causal spatiotemporal attention}, which enforces strict temporal causality and enables efficient frame-by-frame inference via a persistent KV-cache, avoiding re-computation over past frames. Second, we propose \textbf{3D rotary positional embeddings~(3D-RoPE)}, which extends the 2D RoPE to a  \emph{spatiotemporal} relative encoding, allowing the model to reason about where-and-when across long streams.
% \weidi{decide to use capital or small letters for these terms.}
% \yibin{changed to CAP.}

Architecture alone, however, does not guarantee universality: 
the representation needs to be \emph{trained} to encode what downstream tasks need.
To this end, we adopt a unified pre-training recipe that couples three complementary signals, each targeting a different axis of {\em understanding}: 
\textbf{i) static and temporal representation learning}: a self-supervised student-teacher distillation objective that unifies image representation learning with causal video modeling. By distilling both global (image/video-level) and local (patch-level) features, the backbone learns semantic invariances and motion-sensitive dynamics while respecting causality; \textbf{ii) streaming geometric reconstruction}: lightweight feedforward dual DPT and camera heads predict depth maps, ray maps, and camera poses from the stream, injecting explicit 3D constraints so that the representation reflects physical scene structure rather than appearance alone; \textbf{iii) vision-language alignment}: a lightweight auto-regressive language decoder trained on captioning, OCR, and visual grounding connects visual tokens with linguistic concepts, encouraging fine-grained details and semantic grounding that are useful for reasoning-heavy tasks. Together, these objectives encourage \model to learn representations that are simultaneously temporally coherent, geometrically grounded, and language-aligned.

In summary, we propose \textbf{\model}, a unified \emph{streaming} visual backbone that extends a pre-trained image encoder with causal spatiotemporal attention and 3D-RoPE, enabling strictly causal, efficient online inference. Our extensive multi-task pre-training across 29 diverse datasets demonstrate that scaling diverse objectives is highly synergistic rather than merely additive: (i) causal video modeling is essential for capturing dynamic motions; (ii) explicit geometric pre-training is prerequisite for downstream spatial intelligence and embodied control; and (iii) early vision-language alignment within the backbone is critical to prevent catastrophic failures during VLM integration. 
Lastly, we demonstrate that \model serves as a strong \textbf{vision backbone}: its features perform well on standard image/video probing, it natively supports competitive streaming geometric reconstruction, and it enables downstream language models and policy heads to solve general and complex spatial video question answering, and robotic manipulation without backbone fine-tuning, matching or surpassing task-specific baselines. We will release the \textbf{model and training code} to the community.

\section{Related Work}
\vspace{2pt} \noindent \textbf{Vision foundation models.}
The pursuit of a unified vision foundation model capable of generalizing across diverse perceptual dimensions remains a central objective in computer vision. For static images, vision-language contrastive learning frameworks such as CLIP~\cite{clip} and SigLIP~\cite{siglip, siglip2} have demonstrated impressive semantic generalization, while self-supervised paradigms like DINO~\cite{dino, dinov2, dinov3}, MAE~\cite{mae}, and I-JEPA~\cite{ijepa} excel in extracting low-level visual features.
In the video domain, research has pivoted toward capturing complex spatiotemporal dependencies. Supervised models~\cite{vivit, timesformer, streamformer} leverage large-scale video-text corpora, whereas self-supervised approaches focus on reconstructing masked spatiotemporal patches at the pixel level~\cite{videomae, videomae2} or feature level~\cite{vjepa,vjepa2}. 
Despite these advancements, the landscape remains fragmented: most models are either restricted to static perception or rely on non-causal, offline temporal processing. Our \model bridges this gap by unifying image and video representation learning within a strictly causal, streaming framework.

\vspace{2pt} 
\noindent \textbf{Feed-forward 3D reconstruction models.}
DUSt3R~\cite{dust3r} has led a paradigm shift in multi-view geometry, moving from traditional optimization-based methods like Structure-from-Motion (SfM) to learning-based feed-forward neural networks. Building upon this, subsequent works have extended this paradigm to support multi-view and video inputs beyond simple image pairs~\cite{vggt, fast3r, pi3, depthanything3, d4rt}. Parallel to these developments, the research focus has transitioned from offline sequence reconstruction to the more challenging \textit{online} setting~\cite{span3r, cut3r, stream3r, point3r, streamvggt}, which requires real-time geometric reasoning. Although these specialized 3D experts demonstrate remarkable precision in geometric tasks, they often lack high-level semantic reasoning capabilities. 
Our method not only achieves competitive performance in streaming geometric estimation but also exhibits superior generalization across vision tasks that require deep semantic understanding, effectively bridging the gap between spatial structure and semantic abstraction.

\vspace{2pt} 
\noindent \textbf{Visual representation for VLMs and VLAs.}
Advanced Vision-Language Models (VLMs)~\cite{llava,llavanext,llava_one_vision,llavavideo,qwenvl,qwen25vl} rely on robust visual encoders to ground textual reasoning. While contrastive models like CLIP~\cite{clip} and SigLIP~\cite{siglip} serve as the \textit{de facto} standard for global semantics, they typically fall short in fine-grained visual perceptions~\cite{clipblind}. Recent studies have sought to enhance visual representations by adopting more powerful backbones~\cite{siglip2, qwen3vl, llava-ov-1.5}, leveraging ensembles of multiple specialized encoders~\cite{cambrian, deepseek_vl, brave, eagle}, or incorporating specialized geometric experts to bolster spatial understanding~\cite{vlm3r, spacemind}. Building on general VLMs, Vision-Language-Action (VLA) models~\cite{pi0,openvla,rt2} demand both high-level semantics and low-level geometric precision. However, a significant gap exists between current VLMs and the demands of VLA embodied tasks~\cite{robovlm}, especially in visual representation~\cite{vlm4vla}. In contrast, our \model bridges this gap by unifying semantic, dynamic and geometric representations into a single, efficient streaming backbone.
\section{Method}
This section presents \textbf{\model}, a unified \emph{streaming} visual backbone. We formalize causal streaming perception in Sec.~\ref{subsec:problem_formulation}, describe the backbone that turns a pre-trained image ViT into an efficient online encoder in Sec.~\ref{subsec:backbone}, introduce our multi-task training framework in Sec.~\ref{subsec:training}, and detail how the learned representation supports unified perception, reasoning, and action in Sec.~\ref{subsec:downstream}.

\subsection{Problem Formulation}
\label{subsec:problem_formulation}

We aim to train a \textbf{unified streaming visual backbone} that produces a general-purpose representation at each time step of a video stream and can be reused by diverse downstream tasks with minimal task-specific modeling.

Formally, let a continuous video stream be $\mathcal{V}^T=\{\mathcal{I}_1,\mathcal{I}_2,\dots,\mathcal{I}_T\}$, where $\mathcal{I}_t \in \mathbb{R}^{H\times W\times 3}$.
{At time step $t$, the backbone processes the current frame and historical context to produce a composite output state $\mathbf{O}_t$:
\begin{equation}
    \mathbf{O}_t = f_\theta\!\left(\mathcal{I}_t \mid \mathcal{I}_1, \dots, \mathcal{I}_{t-1}\right), \quad t \in \{1,\dots,T\}.
\end{equation}}

To support real-time deployment, we enforce a \textbf{strict causality constraint}: $\mathbf{O}_t$ must not depend on future frames $\{\mathcal{I}_{t+1},\dots,\mathcal{I}_T\}$.
The same backbone is expected to support multiple regimes:
\begin{enumerate}[label=\roman*)]
    \item \textbf{Static perception} ($T=1$): the stream degenerates to a single image; the model acts as an image encoder capturing semantics and layout;
    \item \textbf{Dynamic understanding} ($T>1$): the backbone captures temporal evolution and motion cues from a continuous stream;
    \item \textbf{Geometric reasoning} ($T>1$): by aggregating multi-frame evidence causally, the backbone supports online reconstruction of 3D/4D scene structure;
    \item \textbf{Embodied control} ($T \ge 1$): the backbone supplies real-time, action-oriented representation to drive closed-loop policies for robotic manipulation.
\end{enumerate}

\subsection{Streaming Visual Backbone}
\label{subsec:backbone}
\begin{figure}[t]
    \centering
    \includegraphics[width=1.0\textwidth]{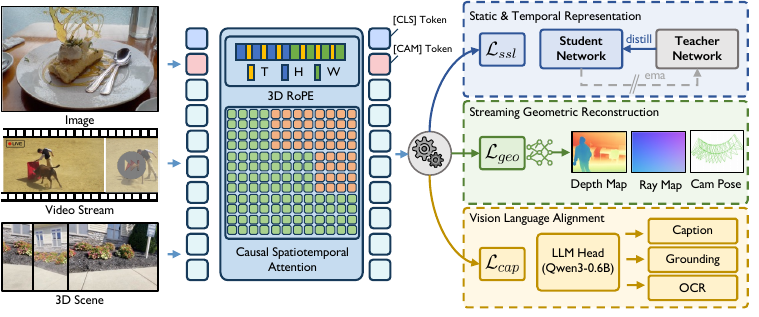}
    \caption{\textbf{Overall framework of \model.} Equipped with 3D-RoPE and causal spatiotemporal attention, our unified backbone is trained via a multi-task framework that couples static and temporal representation learning, streaming geometric reconstruction, and vision-language alignment.}
    \label{fig:arch}
    \vspace{-1em}
\end{figure}

\noindent \textbf{Overview.}
We instantiate \textbf{\model} as a vision transformer from DINOv3~\cite{dinov3}, leveraging its strong spatial representation as a starting point. As depicted in Fig.~\ref{fig:arch}, to evolve an image encoder into a \textbf{unified streaming} backbone, we introduce two modifications: 
(i) \textbf{causal spatiotemporal attention} to enable online inference with temporal causality and a persistent KV-cache; (ii) \textbf{3D rotary positional embeddings (3D-RoPE)}, extending the 2D positional priors to the spatiotemporal domain. 
These changes allow efficient processing of long streams while preserving the pre-trained spatial inductive bias.

\vspace{2pt} \noindent \textbf{Tokenization and outputs.}
For a video stream $\mathcal{V}^T \in \mathbb{R}^{T\times H\times W\times 3}$, we split each frame into non-overlapping $p\times p$ patches, producing $h\times w$ patches per frame with $h=H/p$ and $w=W/p$. After linear projection, we prepend per-frame special tokens: one global \texttt{[CLS]} token, four register tokens~\cite{vision_registers}, and an optional \texttt{[CAM]} token. Let \(N_s\) be the number of special tokens per frame; the input sequence is $\mathbf{z}^{0} \in \mathbb{R}^{T \times (N_s + hw) \times d}$.

After multiple Transformer blocks, we extract (i) dense spatiotemporal feature maps from selected intermediate layers $\mathbf{L}$, and (ii) final-layer special tokens for global semantics and camera prediction:
\begin{equation}
    \mathbf{O} = \left\{ \mathcal{Z}, \mathbf{z}_{\text{cls}}, \mathbf{z}_{\text{cam}} \right\} = f_\theta\left( \mathbf{z}^0 \right),
\end{equation}
where $\mathcal{Z}$ denotes selected layers' dense patch feature \( \left\{ \mathbf{z}^{(\ell)} \in \mathbb{R}^{T\times h\times w\times d} \right\}_{\ell \in \mathbf{L}} \), the last-layer $\texttt{[CLS]}$ token $\mathbf{z}_{\text{cls}} \in \mathbb{R}^{T \times d}$ provides an image/video-level summary, and the $\texttt{[CAM]}$ token $\mathbf{z}_{\text{cam}} \in \mathbb{R}^{T \times d}$ is routed to the camera head for pose estimation.

\vspace{2pt} \noindent \textbf{Causal spatiotemporal attention.}
A naive Transformer that attends over all frames violates causality and is inefficient for streaming. We therefore apply spatiotemporal self-attention with a \textbf{causal temporal mask}, so tokens at time $t$ may only attend to tokens from times $\le t$.

Let the total number of tokens be $N_{\text{total}} = T \cdot (N_s + hw)$, the attention mask as $\mathbf{M} \in \{0,-\infty\}^{N_{\text{total}}\times N_{\text{total}}}$.
The masked attention is formalized as: 
\begin{equation}
    \mathrm{Attn}(\mathbf{Q},\mathbf{K},\mathbf{V})
    = \mathrm{Softmax}\!\left(\frac{\mathbf{Q}\mathbf{K}^\top}{\sqrt{d_{head}}} + \mathbf{M}\right)\mathbf{V},
\end{equation}

For a query token with global index $u$ and a key token with global index $v$, the mask is defined via a frame-index mapping $\tau(\cdot)$:
\begin{equation}
    M_{u,v} =
    \begin{cases}
    0, & \text{if } \tau(u) \ge \tau(v)\\
    -\infty, & \text{if } \tau(u) < \tau(v)
    \end{cases},
\end{equation}
where $\tau(u)$ returns the time step of token $u$.
This ensures every patch in frame $t$ can attend to all patches in frames $\le t$, but never to future frames.

\paragraph{Streaming inference with KV-cache.}
Under the above causality, inference can be performed incrementally: when frame $t$ arrives, we compute its queries and reuse cached keys/values from frames $\le t-1$. This yields online processing without recomputing attention over the full past stream, while producing the same output as full causal attention.

\vspace{2pt} 
\noindent \textbf{3D rotary positional embeddings~(3D-RoPE).}
The original DINOv3~\cite{dinov3} uses 2D RoPE to encode relative spatial offsets.
We extend RoPE to the spatiotemporal domain by re-purposing the per-head feature dimension $d_{\text{head}}$ while preserving as much pre-trained spatial structure as possible. Concretely, we adopt an allocation strategy with a \textbf{2:3:3} split across $(t,y,x)$, namely, time, height, and width. We interleave temporal components into the original 2D RoPE such that indices \(i \equiv 3 \pmod 4\) encode time \(t\)~\cite{qwen3vl}, while the remaining indices follow the original DINOv3 pattern for \(y\) and \(x\). We further apply RoPE-box jittering~\cite{dinov3} to improve robustness of the positional embeddings.

\subsection{Unified Multi-task Learning Objectives} 
\label{subsec:training}
A unified streaming backbone should ideally produce representations that transfer seamlessly across semantics, dynamics, geometry, and language.
We therefore train \model with a unified multi-task framework that combines three complementary objectives:
\textbf{static \& temporal representation learning}, \textbf{streaming geometric reconstruction}, and \textbf{vision-language alignment}. These signals are designed to be compatible with causal streaming environments and to encourage representations that are simultaneously discriminative, temporally coherent, and physically grounded.

\vspace{2pt} \noindent \textbf{Static \& temporal representation learning.} We treat an image as a degenerate stream with \(T=1\), enabling a single objective to cover both images and videos.
We employ a DINOv3-style student-teacher distillation objective~\cite{dinov3}:
\begin{equation}
    \mathcal{L}_{\mathrm{ssl}} =
    \mathcal{L}_{\mathrm{DINO}} +
    \mathcal{L}_{\mathrm{iBOT}} +
    0.1 \times \mathcal{L}_{\mathrm{KoLeo}} + \mathcal{L}_{\mathrm{gram}},
\end{equation}
Given a set of views (global/local crops from an image or a video clip), the four terms encourage (i) global semantic consistency~($\mathcal{L}_{\text{DINO}}$)~\cite{dino}, (ii) patch-level discriminative features~($\mathcal{L}_{\text{iBOT}}$)~\cite{ibot}, (iii) a well-spread feature space~($\mathcal{L}_{\text{KoLeo}}$)~\cite{koleo},  and (iv) a consistent patch-level feature during training
~($\mathcal{L}_{\text{gram}}$)~\cite{dinov3}.  
We provide full formulations in the appendix.

\vspace{2pt} \noindent \textbf{Streaming geometric reconstruction.}
To inject explicit 3D constraints, we attach feed-forward geometric heads that reconstruct monocular streams into a coherent dynamic representation. We use: (i) a depth head implemented as a dual-DPT module~\cite{depthanything3} to predict dense depth maps and ray maps; and (ii) a lightweight MLP camera head to predict camera poses.

Using the selected backbone features $\{\mathbf{z}^{(\ell)} \in \mathbb{R}^{T\times h\times w\times d}\}_{\ell \in \mathbf{L}}$, the depth head predicts depth maps $\hat{D} \in \mathbb{R}^{T\times H\times W\times 1}$ and ray maps $\hat{R} \in \mathbb{R}^{T\times H\times W\times 6}$. The camera head consumes camera tokens $\mathbf{z}_{\text{cam}}$ and predicts per-frame pose parameters $\hat{g}\in \mathbb{R}^{T\times 9}$, which is the concatenation of the rotation quaternion $\mathbf{q} \in \mathbb{R}^{T \times4}$, the translation vector $\mathbf{t} \in \mathbb{R}^{T \times 3}$, and the field of view $\mathbf{f} \in \mathbb{R}^{T \times 2}$, with the camera's principal point assumed to be at the image center. Following~\cite{vggt}, we supervise these outputs against normalized ground-truth depth maps ${D}$, ray maps ${R}$, point maps ${P}$, and camera poses ${g}$ and perform a multi-task geometric loss:
\begin{equation}
    \mathcal{L}_{\mathrm{geo}} =
    \mathcal{L}_{\mathrm{depth}} +
    \mathcal{L}_{\mathrm{ray}} +
    \mathcal{L}_{\mathrm{points}} +
    \mathcal{L}_{\mathrm{camera}},
\end{equation}

\paragraph{Depth with confidence.}
We use an L1 regression loss with an additional gradient term and a learned confidence \(c_{t}\):
\begin{equation}
    \mathcal{L}_{\text{depth}} =
    \sum_{t=1}^{T}
    \left(
    \left\| c_{t} (\hat{D}_{t} - D_{t}) \right\|_1
    + \left\| c_{t} (\nabla \hat{D}_{t} - \nabla D_{t}) \right\|_1
    - \alpha \log c_{t}
    \right),
\end{equation}

where \(\nabla\) is the spatial gradient operator and \(\alpha\) is a hyper-parameter.

\paragraph{Rays, points, and camera.}
Similarly, we apply standard L1 regression to ray maps, point maps, and camera poses:
\begin{equation}
    \mathcal{L}_{\text{ray}} =
    \sum_{t=1}^{T} \left\| \hat{R}_{t} - R_{t} \right\|_1,
    \quad
    \mathcal{L}_{\text{points}} =
    \sum_{t=1}^{T} \left\| \hat{P}_{t} - P_{t} \right\|_1,
    \quad
    \mathcal{L}_{\text{camera}} = \sum_{t=1}^{T} \left\| \hat{g}_t - g_t \right\|_1
\end{equation}
where the camera ray is defined by concatenating the ray origin $\mathbf{o} \in \mathbb{R}^3$ and direction $\mathbf{d} \in \mathbb{R}^3$. And point maps can thus be derived as $\hat{P}_{t} = \hat{\mathbf{o}}_{t} + \hat{D}_{t}\odot \hat{\mathbf{d}}_{t}$. 

\vspace{2pt} 
\noindent \textbf{Vision-language alignment.}
To align visual tokens with linguistic concepts and enable reasoning-oriented downstream tasks, we attach an MLP projector and lightweight autoregressive language decoder (Qwen3-0.6B~\cite{qwen3}) to the vision backbone.
This branch is trained on a mixture of vision-language tasks including dense captioning, OCR, and object grounding.

Given the visual tokens from the last layer \(\mathbf{z} ^ L\) and an instruction prompt \(\mathbf{x}_{\text{inst}}\), we train with the standard language modeling objective:
\begin{equation}
    \mathcal{L}_{\text{cap}} =
    -\sum_{n=1}^{L_{\text{text}}}
    \log P_{\text{text}}\!\left(y_n \mid \mathbf{z} ^ L, \mathbf{x}_{\text{inst}}, \mathbf{y}_{<n}\right),
\end{equation}
where \(P_{\text{text}}\) is the token distribution predicted by the decoder and \(\mathbf{y}_{<n}\) denotes previously generated tokens.
During training, gradients are back-propagated through the language decoder into the vision backbone, injecting fine-grained semantic and spatial supervision, which proves to be essential capabilities for VLM and VLA integration in our experiments.

\vspace{2pt} 
\noindent \textbf{Total objective.}
The overall training loss is a weighted combination:
\begin{equation}
    \mathcal{L}_{\text{total}} =
    \lambda_{\mathrm{ssl}} \cdot \mathcal{L}_{\mathrm{ssl}} +
    \lambda_{\mathrm{geo}} \cdot \mathcal{L}_{\mathrm{geo}} +
    \lambda_{\mathrm{cap}} \cdot \mathcal{L}_{\mathrm{cap}},
\end{equation}
where we set $\lambda_{\mathrm{ssl}} = 0.1$ and $\lambda_{\mathrm{geo}} = \lambda_{\mathrm{cap}} = 1$ to balance the loss magnitudes.

\subsection{Universal Representation for Downstream Applications}
\label{subsec:downstream}
A foundation vision backbone should be reusable without expensive full-model adaptation. Accordingly, we evaluate \model in a \textbf{strictly frozen} setting, where the backbone parameters are kept fixed and only specific task modules are trained. We organize downstream tasks into three increasing levels of complexity:

\vspace{2pt} \noindent \textbf{Perception: image \& video.} 
For discriminative perception tasks, we attach lightweight heads on top of frozen vision features. For static image tasks ({\em e.g.}, segmentation), we apply linear decoders on spatial tokens; for video tasks, we use attentive pooling modules on the frozen spatiotemporal features.

\vspace{2pt} \noindent \textbf{Reasoning: vision-language models~(VLM).}
To enable spatial and temporal reasoning, we integrate \model with an LLM by using an MLP projector that maps frozen visual tokens into the language embedding space. The LLM then generates text conditioned on the projected visual context and textual instructions. Throughout, the vision backbone remains frozen, forcing the LLM to interpret the intrinsic visual representation.

\vspace{2pt} \noindent \textbf{Action: vision-language-action~(VLA).}
For embodied control, we extend the VLM setup by attaching an MLP-based action expert to the LLM outputs to predict robot actions $a_t$ from the frozen visual observations and language instructions.
Crucially, because our pre-training objectives ($\mathcal{L}_{\mathrm{ssl}}, \mathcal{L}_{\mathrm{geo}}$) explicitly encode dynamics and geometry, the frozen features provide a strong bridge from perception to action without domain-specific visual fine-tuning.

\section{Experiments}
\subsection{Pre-training Data}
We train our model on a wide range of datasets featuring static images, dynamic videos, and geometric 3D/4D Scenes. 
Totally, our training pipeline consumes approximately 200M frames from 29 datasets, providing a dense and diverse supervisory signal that ensures the model's robustness across various streaming scenarios. A detailed breakdown of the pre-training data is provided in Tab.~\ref{tab:datasets}.

\begin{table}[t]
\centering
\small
\begin{tabularx}{\linewidth}{l >{\raggedright\arraybackslash}X r}
\toprule
\textbf{Task} & \textbf{Pre-training Dataset} & \textbf{Frames} \\
\midrule
 \textbf{Image SSL} & DataComp-100M~\cite{datacomp}, ImageNet-21K~\cite{imagenet21k} & $\sim$113M \\ 
 
 \addlinespace[3pt]
 
 \textbf{Video SSL} & Kinetics~\cite{k400}, SSv2~\cite{ssv2}, PE-Videos~\cite{PerceptionEncoder} & $\sim$20M \\

 \addlinespace[3pt]
 
 \textbf{3D/4D Scenes} & Co3Dv2~\cite{co3d}, WildRGBD~\cite{wildrgbd}, BlendedMVS~\cite{BlendedMVS}, MegaDepth~\cite{megadepth}, MVS-Synth~\cite{mvssynth}, PointOdyssey~\cite{pointodyssey}, HyperSim~\cite{hypersim}, TartanAir~\cite{tartanair}, DL3DV~\cite{dl3dv}, ScanNet~\cite{scannet}, ScanNet++~\cite{scannetpp}, ArkitScenes~\cite{arkitscenes}, Virtual Kitti 2~\cite{vkitti2}, MapFree~\cite{mapfree}, Dynamic Replica~\cite{dynamicreplica}, Spring~\cite{spring}, OmniWorld-Games~\cite{omniworld},  Aria Synthetic Environments~\cite{ariasynenv}, Waymo-Open~\cite{waymo} & $\sim$18M \\
  \addlinespace[3pt]
 \textbf{Captioning} & GRIT~\cite{Kosmos2, coyo-700m}, RefCOCO Series~\cite{refcoco}, Blip3-OCR~\cite{blip3}, SA1B-Caption~\cite{chen2024far} & $\sim$50M \\
 \midrule
 \textbf{Total} & - & $\sim$200M \\
\bottomrule
\end{tabularx}
\caption{\textbf{Overview of pre-training datasets.} We leverage a diverse collection of images, videos, and 3D/4D data.}
\label{tab:datasets}
% \vspace{-1em}
\end{table}

\subsection{Implementation Details}
We initialize our model using the pre-trained weights of DINOv3~\cite{dinov3} ViT-L. 
We employ a multi-task learning strategy combined with gradient accumulation, where we sequentially interleave all tasks within each training step and perform a single model parameter update after traversing the batches from all tasks. 
The training is conducted on 64$\times$ NVIDIA H200 GPUs and is divided into two stages: the first trained for 60K steps at a resolution of $224 \times 224$, followed by a second stage trained for 120K steps at a resolution of $512 \times 512$. For optimization, we use Adam optimizer~\cite{adam} with a peak learning rate of $1 \times 10^{-4}$, employing a 4K-step warmup followed by a cosine annealing schedule. Weight decay is disabled in both stages. All multi-frame sequences, including video and geometric scenes, are sampled with a sequence length of $T=16$ frames per sample. 
Benefiting from our causal mask design, each training sample effectively provides distinct supervision signals for varying temporal contexts ranging from 1 to $T$ frames.

\subsection{Downstream Evaluations}
To validate the universality and robustness of our learned representations, we structure our downstream experiments into four phases, as illustrated in Tab.~\ref{tab:holistic_eval}:
(i) \textbf{image \& video probing}, which assesses the quality of global and dense features on standard image and video benchmarks;
(ii) \textbf{streaming geometric reconstruction}, which evaluates the model's capability to estimate continuous camera poses and depth maps in an online fashion;
(iii) \textbf{visual backbone for VLMs}, 
which incorporates the representation into Vision-Language Models (VLMs) for complex video and spatial reasoning; and
(iv) \textbf{visual backbone for VLA policies}, where the model serves as the visual engine for Vision-Language-Action (VLA) policies in robotic manipulation. 
Crucially, to demonstrate its capacity as a truly universal foundational vision model, \textbf{we keep the backbone strictly frozen} across all downstream tasks.

\begin{table}[t]
    \centering
    \small
    \vspace{-1em}
    \setlength{\aboverulesep}{0pt}
    \setlength{\belowrulesep}{0pt}
    \renewcommand{\arraystretch}{1.2}
    \setlength{\tabcolsep}{9.2pt}
    % \resizebox{\textwidth}{!}{
    \begin{tabular}{l l |  >{\columncolor{highlightcolor}}c  c c c  c c }
        \toprule 
        & {\bf Benchmark} &
        {\bf Ours} & 
        {\bf DINOv3-L} & 
        {\bf V-JEPA2-L} & 
        {\bf CUT3R} & 
        {\bf LLaVA-Video} & 
        {\bf OpenVLA} \\
        \midrule
        \multirow{3}{*}{\bf\it Image} 
        &\textbf{ImageNet}$_{\text{cls}}$$\uparrow$ & 84.7 & 86.7 & - & - & - & - \\
        &\textbf{NYUv2}$_{\text{monodepth}}$$\downarrow$ & 0.377 & 0.377 & - & - & - & -  \\
        & \textbf{ADE20K}$_{\text{seg}}$$\uparrow$ & 49.1 & 51.5 & - & - & - & - \\
        \midrule
        \multirow{3}{*}{\bf\it Video} 
        & \textbf{SSv2}$_{\text{act}}\uparrow$ & 68.5 & 54.0 & 73.7 & - & - & -  \\
        & \textbf{K400}$_{\text{act}}\uparrow$ & 85.7 & 83.6 & 85.1 & - & - & -  \\
        & \textbf{DAVIS'17}$_{\text{vos}}\uparrow$ & 71.6 & 73.2 & 44.2 & - & - & - \\
        \midrule
        \multirow{6}{*}{\shortstack{\bf\it 3D \\ \bf\it Geom.}}
        & \textbf{Sintel}$_{\text{videodepth}}$ $\downarrow$ & 0.314 & - & - & 0.421 & - & - \\
        & \textbf{BONN}$_{\text{videodepth}}\downarrow$ & 0.072 & - & - & 0.078 & - & - \\
        & \textbf{KITTI}$_{\text{videodepth}}\downarrow$ & 0.136 & - & - & 0.118 & - & - \\
        \cmidrule{2-8} 
        & \textbf{Sintel}$_{\text{pose}}\downarrow$ & 0.227 & - & - & 0.213 & - & -  \\
        & \textbf{TUM}$_{\text{pose}}\downarrow$ & 0.049 & - & - & 0.046 & - & - \\
        & \textbf{ScanNet}$_{\text{pose}}\downarrow$ & 0.076 & - & - & 0.099 & - & - \\
        \midrule
        \multirow{5}{*}{\bf\it VLM} 
        & \textbf{VideoMME}$_{\text{vqa}}\uparrow$ & 60.7 & - & - & - & 61.8 & - \\
        & \textbf{VideoMMMU}$_{\text{vqa}}\uparrow$ & 40.0 & - & - & - & 38.7 & - \\
        & \textbf{PerceptionTest}$_{\text{vqa}}\uparrow$ & 68.9 & - & - & - & 67.6 & - \\
        & \textbf{EgoSchema}$_{\text{vqa}}\uparrow$ & 60.9 & - & - & - & 57.3 & - \\
        & \textbf{VSI-Bench}$_{\text{vqa}}\uparrow$ & 70.6 & - & - & - & 35.6 & - \\
        \midrule
        \multirow{2}{*}{\bf\it VLA} 
        & \textbf{CALVIN}$_{\text{mani}}\uparrow$ & 3.89 & - & - & - & - & 2.55 \\
        & \textbf{Simpler-Bridge}$_{\text{mani}}\uparrow$ & 45.8 & - & - & - & - & 53.7 \\
        \bottomrule
    \end{tabular}
    % }
\caption{
{\bf Holistic evaluation.} 
We compare \model across 5 domains. Subscripts denote the specific task (\emph{i.e.}, cls: classification, monodepth: monocular depth estimation, seg: semantic segmentation, act: video action recognition, vos: video object segmentation, videodepth: video depth estimation, pose: camera pose estimation, vqa: visual question answering, mani: robotic manipulation). 
The evaluated metrics are ACC@1 for cls and act, RMSE for monodepth, mIOU for seg, $\mathcal{J}\&\mathcal{F}$ for vos, absRel for videodepth, ATE for pose, Accuracy (Acc.) for vqa, Average Length (Avg. Len) for CALVIN, and Success Rate (SR \%) for Simpler-Bridge. 
Here, ``-'' indicates that a specialized baseline is not natively applicable to or lacks the capability for the given task.
}
\label{tab:holistic_eval}
% \vspace{-1em}
% \vspace{-6mm}
\end{table}

\vspace{2pt}
\noindent \textbf{Image \& video probing.}
We first evaluate the versatility of the frozen features extracted by \model across both static and dynamic domains. 
Following the DINOv3~\cite{dinov3} protocol for image tasks, we train task-specific linear heads on top of the frozen backbone to evaluate ImageNet-1K~\cite{imagenet} classification, ADE20K~\cite{ade20k} semantic segmentation, and NYUv2~\cite{NYUv2} monocular depth estimation. 
For video understanding, we probe action recognition on Kinetics-400~(K400)~\cite{k400} and Something-Something V2~(SSv2)~\cite{ssv2} using attentive pooling from the V-JEPA 2~\cite{vjepa2} setup. 
Furthermore, we evaluate dense spatiotemporal tracking on the DAVIS'17~\cite{davis17} Video Object Segmentation (VOS) benchmark.

As presented in Tab.~\ref{tab:holistic_eval}, \model successfully bridges the gap between static and dynamic perception.
On image benchmarks, it preserves robust spatial discrimination, achieving competitive results on dense prediction tasks (NYUv2 and ADE20K) comparable to the image specialist DINOv3. In contrast, \model demonstrates superior temporal understanding ability, significantly outperforming DINOv3 on the motion-intensive SSv2 dataset (68.5\% vs. 54.0\%).
{Unlike typical video backbones that exhibit poor spatial alignment (\emph{e.g.}, V-JEPA 2 at 44.2 $\mathcal J\&F$),  \model leverages the KV-cache to enable a long-range extraction of the entire video. Our approach yields a $\mathcal J\&F$ mean of 71.6, performing on par with leading image-based per-frame extractors (DINOv3 at 73.2).}
This confirms that our unified multi-task training effectively injects temporal motion dynamics into the representations, without compromising the fine-grained spatial priors required for dense image tasks.

\vspace{2pt} \noindent \textbf{Streaming geometric reconstruction.} To evaluate the geometric ability of our streaming representations, we benchmark \model on online 3D reconstruction tasks following CUT3R~\cite{cut3r}. 
We assess online video depth estimation across Sintel~\cite{sintel}, BONN~\cite{BONN}, and KITTI~\cite{kitti} (Tab.~\ref{tab:video_depth}), and online camera pose estimation on Sintel, TUM Dynamics~\cite{tumd}, and ScanNet~\cite{scannet} (Tab.~\ref{tab:camera_pose}). 

Across both tasks, our unified encoder consistently establishes highly competitive or superior results against specialized online 3D models, validating its robustness in real-time spatial perception. Notably, benefiting from our causal temporal attention mechanism, the architecture natively supports KV-caching during inference. 
This allows the model to process incoming video streams frame-by-frame with $O(T)$ temporal complexity per step, avoiding the redundant re-computation typical of bidirectional global attention. 
Furthermore, by leveraging the geometric extrapolation properties inherent to our 3D-RoPE, \model can seamlessly generalize to much longer, unseen sequences that significantly exceed the training horizon. 
For instance, although \model is pre-trained with a fixed temporal window of only $T=16$ frames, it demonstrates impressive zero-shot length extrapolation capabilities, effortlessly processing continuous streams of up to 110 frames during evaluation.
\begin{table}[t]
\setlength{\tabcolsep}{7pt}
\centering
  \small
  % \resizebox{\textwidth}{!}{
  \begin{tabular}{l c | c c | c c |c c}
  \toprule
 && \multicolumn{2}{c|}{\bf Sintel}& \multicolumn{2}{c|}{\bf BONN}& \multicolumn{2}{c}{\bf KITTI}\\
\bf Method & \bf Param. & Abs Rel$\downarrow$ & {$\delta < 1.25$}$\uparrow$ & Abs Rel$\downarrow$ & $\delta < 1.25$$\uparrow$& Abs Rel$\downarrow$ & $\delta < 1.25$$\uparrow$\\
\midrule
Span3R~\cite{span3r} & 600M &        0.622  & 42.6 & 0.144 & 81.3 & 0.198 & 73.7 \\
Cut3R~\cite{cut3r} & 600M &         \underline{0.421}  & \underline{47.9} &0.078 & 93.7 & \underline{0.118} & \underline{88.1}\\
Point3R~\cite{point3r} & 600M & 0.481 & 44.8 & \textbf{0.066} & \underline{95.8} & \textbf{0.093} & \textbf{93.5} \\ 
\midrule
\bf Ours & 400M &                                                           \textbf{{0.314}}   & \textbf{58.3} &\underline{0.072} &\textbf{96.7} & 0.136& 81.8 \\
\bottomrule
\end{tabular}
\caption{{\bf Comparison on online video depth estimation.}}
\label{tab:video_depth}
\end{table}

\begin{table}[t]
    \centering
      \small
      \resizebox{\textwidth}{!}{
      \begin{tabular}{l c |c c c |c c c |c c c}
      \toprule
     && \multicolumn{3}{c|}{\bf Sintel}& \multicolumn{3}{c|}{\bf TUM-dynamics}& \multicolumn{3}{c}{\bf ScanNet}\\
\bf Method & \bf Param. & ATE$\downarrow$ & RPE$_{T}\downarrow$ & RPE$_{R}\downarrow$ & ATE$\downarrow$ & RPE$_{T}\downarrow$ & RPE$_{R}\downarrow$ & ATE$\downarrow$ & RPE$_{T}\downarrow$ & RPE$_{R}\downarrow$\\
\midrule
Span3R~\cite{span3r} & 600M &       0.329  & 0.110  & 4.471 & 0.056 & 0.021 & \underline{0.591} & \underline{0.096} & 0.023 & 0.661 \\
Cut3R~\cite{cut3r} & 600M &        \textbf{0.213}  & \underline{0.066}  & \textbf{0.621} & \textbf{0.046} & \underline{0.015} & \textbf{0.473} & 0.099 & \underline{0.022} & \textbf{0.600} \\
Point3R~\cite{point3r} & 600M & 0.442 & 0.154 & 1.897 & 0.058 & 0.031 & 0.758 & 0.097 & 0.035 & 2.791 \\
\midrule
\bf Ours  & 400M &                                                           \underline{0.227}& \textbf{{0.051 }}& \underline{0.695} & \underline{0.049} & \textbf{ 0.012} &  0.601 &\textbf{ 0.076} &\textbf{ 0.017} &  \underline{0.659} \\
\bottomrule
\end{tabular}
}
\caption{{\bf Comparison on online video camera pose estimation.}}
\label{tab:camera_pose}
% \vspace{-1em}
\end{table}

\noindent \textbf{Visual backbone for VLMs.}
Adhering to the framework of LLaVA-Video~\cite{llavavideo}, we evaluate the efficacy of \model as a general-purpose visual encoder for Video LLMs. 
We integrate our backbone with Qwen2.5-7B-Instruct~\cite{qwen25} to perform general and spatial video question answering. 
Further training details are provided in the {appendix}.

As shown in the \textit{VLM} section of Tab.~\ref{tab:holistic_eval}, despite a frozen visual encoder and an identical general video data recipe, \model achieves better performance than LLaVA-Video on general Video QA benchmarks.
Although marginally behind on VideoMME~\cite{videomme}, it consistently achieves superior results across other major benchmarks like VideoMMMU~\cite{videommmu}, PerceptionTest~\cite{perceptiontest}, and EgoSchema~\cite{egoschema}, highlighting its effectiveness.
Our model further demonstrates exceptional spatial reasoning capabilities. As shown in Tab.~\ref{tab:vsi_bench}, on the \textbf{VSI-Bench}~\cite{vsibench}, a challenging benchmark for spatial intelligence, our model achieves state-of-the-art performance. 
Remarkably, it surpasses specialized geometry-aware baselines equipping extra geometry encoders, such as VLM-3R~\cite{vlm3r} and SpaceMind~\cite{spacemind}.
This result strongly validates that our unified streaming pre-training naturally engenders rich, actionable spatial understanding within the visual features themselves, eliminating the need for auxiliary geometric modules. 
\begin{table}[t]
    \centering
    % \vspace{-1em}
    \small
    \begin{tabular}{l|c|ccccccccc}\toprule
    \textbf{Methods} &\textbf{Avg.} &\textbf{\makecell{Obj.\\Count}} &\textbf{\makecell{Abs.\\Dist.}} &\textbf{\makecell{Obj.\\Size}} &\textbf{\makecell{Room\\Size}} &\textbf{\makecell{Rel.\\Dist}} &\textbf{\makecell{Rel.\\Dir.}} &\textbf{\makecell{Route\\Plan}} &\textbf{\makecell{Appr.\\Order}} \\\cmidrule{1-10}
    GPT-4o~\cite{gpt4o} &34.0 &46.2 &5.3 &43.8 &38.2 &37.0 &41.3 &31.5 &28.5 \\
    Gemini-1.5-Pro~\cite{gemini_1.5} &45.4 &56.2 &30.9 &64.1 &43.6 &51.3 &46.3 &36.0 &34.6 \\\midrule
    LLaVA-OneVision-7B~\cite{llava_one_vision} &32.4 &47.7 &20.2 &47.4 &12.3 &42.5 &35.2 &29.4 &24.4 \\
    LLaVA-Video-7B~\cite{llavavideo} &35.6 &48.5 &14.0 &47.8 &24.2 &43.5 &42.4 &34.0 &30.6 \\
    Qwen2.5-VL-7B~\cite{qwen25vl} &32.7 &34.5 &19.4 &47.6 &40.8 &32.8 &24.5 &32.5 &29.4 \\
    % SAT-7B~\cite{sat} &- &- &- &- &- &47.3 &41.1 &37.1 &36.1 \\
    % InternVL-Spatial-8B~\cite{internspatial} &- &68.7 &40.9 &63.1 &54.3 &47.7 &- &29.9 &60.5 \\
    SpaceR-7B~\cite{spacer} &43.5 &61.9 &28.6 &60.9 &35.2 &38.2 &46.0 &31.4 &45.6 \\
    VILASR-7B~\cite{VILASR} &45.4 &63.5 &34.4 &60.6 &30.9 &48.9 &45.2 &30.4 &49.2 \\
    VLM-3R-7B~\cite{vlm3r} &60.9 &70.2 &{49.4} &69.2 &67.1 &{65.4} &{80.5} &\textbf{45.4} &40.1 \\
    % VST-3B ~\cite{vst} & 57.9 & 69.3 & 45.4 & 71.8 & 62.4 &59.0 & 46.0 & 38.7 & 70.2 \\
    VST-7B ~\cite{vst} & 60.6 & 72.0 & 44.4 & 74.3 & 68.3 & 59.7 & 55.8 & \underline{44.9} & 65.2 \\
    SpaceMind~\cite{spacemind} & \underline{69.6} & \textbf{73.3} & \textbf{61.4} & \textbf{77.3} & \underline{74.2} & \underline{67.2} & \textbf{88.4} & 44.3 & \underline{70.5} \\
    \midrule
    \model-7B & \textbf{70.6} & \underline{73.2} & \underline{55.7} & \underline{76.9} & \textbf{74.8} & \textbf{72.3} & \underline{82.1} & \textbf{45.4} & \textbf{84.6}\\ 
    \bottomrule
    \end{tabular}
    \caption{\textbf{Comparison on VSI-Bench.}}
    \label{tab:vsi_bench}
\end{table}

\vspace{2pt} \noindent \textbf{Visual backbone for VLA policies.}
Finally, we extend our evaluation to Embodied AI, assessing \model as the ``perception brain'' for VLA policies. Following VLM4VLA~\cite{vlm4vla}, we adapt \model-7B into a Vision-Language-Action model by attaching a lightweight MLP action head, trained on two simulated robotic manipulation benchmarks: \textbf{CALVIN} (ABC-D split)\cite{calvin} for long-horizon instruction following, and \textbf{SIMPLER-ENV} (Bridge V2 dataset)\cite{simplerenv} for real-to-sim generalization.

Crucially, the visual encoder remains \textbf{fully frozen} during adaptation. As shown in Tab.~\ref{tab:calvin} and Tab.~\ref{tab:bridge}, strong general-purpose VLMs like Qwen2.5-VL struggle significantly in this setting, failing to translate high-level semantic knowledge into precise low-level control signals, highlighting a critical \textit{misalignment} between general visual features and the requirements of embodied interaction, such as depth perception and dynamics prediction. In contrast, \model achieves strong performance even with a frozen backbone, reaching success rates of 3.885 on CALVIN and 45.8\% on SIMPLER-ENV. This underscores the universality of our representation: by explicitly encoding 3D geometry and temporal dynamics during pre-training, \model bridges the gap between perception and action. To the best of our knowledge, this is the \textbf{first visual encoder} to demonstrate such effective zero-shot transferability to VLA benchmarks without domain-specific visual fine-tuning, paving the way for more efficient and generalizable embodied agents.

\begin{table}[t]

\setlength{\tabcolsep}{3.5mm}
\centering
\small
% \begin{tabular}{r r | r r r r r | r}
\begin{tabular}{l  | c c c c c | c}
    % {p{3.2cm} p{0.9cm} p{4.2cm} r r r r r r}
    \toprule
    \textbf{Model (VLM Backbone)}& \textbf{Task-1 }& \textbf{Task-2} & \textbf{Task-3} & \textbf{Task-4} &\textbf{ Task-5} & \textbf{Calvin$\uparrow$} \\
    \midrule
      \skyblue\multicolumn{7}{c}{{Expert Vision-Language-Action Models}}\\
      \midrule
    OpenVLA* (Llama-2)~\cite{openvla}& 0.792 & 0.644 & 0.499 & 0.368 & 0.245 & 2.548 \\
    pi0* (Paligemma-1)~\cite{pi0}& 0.896 & 0.785 & 0.786 & 0.610 & 0.532 & 3.509 \\
    \midrule
      \skyblue\multicolumn{7}{c}{{VLM with VLM4VLA Models (Full Fine-tuning)}}\\
    \midrule
    % Qwen2.5VL-3B & 3.8B   & 0.922 & 0.842 & 0.766 & 0.700 & 0.626 & 3.856 \\
    Qwen2.5VL-7B~\cite{qwen25vl}  & 0.935 & 0.864 & 0.807 & 0.758 & 0.693 & 4.057 \\
    % Qwen3VL-2B & 2.1B & 0.943 & 0.882 & 0.831 & 0.776 & 0.710 & 4.142 \\
    % Qwen3VL-4B & 4.4B & 0.933 & 0.857 & 0.790 & 0.719 & 0.644 & 3.943 \\
    Qwen3VL-8B~\cite{qwen3vl}  & 0.940 & 0.868 & 0.797 & 0.746 & 0.684 & 4.035 \\
    % Qwen3VL-30B-A3B & 31.1B & 0.939 & 0.877 & 0.820 & 0.757&0.682 & 4.075 \\
    Paligemma-1-3B~\cite{paligemma}  & 0.914 & 0.813 & 0.692 & 0.599 & 0.488 & 3.506 \\
    Paligemma-2-3B~\cite{paligemma2} & 0.901 & 0.775 & 0.669 & 0.575 & 0.486 & 3.406 \\
    KosMos-2-1.7B~\cite{Kosmos2}  & 0.878 & 0.721 & 0.591 & 0.498 & 0.408 & 3.096 \\
    % InternVL3.5-4B & 4.7B & 0.934 & 0.864 & 0.795 & 0.728 & 0.656 & 3.977 \\
    % \textbf{\model-7B} & 7.9B & 0.937 & 0.863 &0.800 & 0.75 & 0.686& 4.036 \\
    \midrule
      \skyblue\multicolumn{7}{c}{{VLM with VLM4VLA Models (Frozen Vision)}}\\
    \midrule
      Qwen2.5VL-7B & 0.901 & 0.700 & 0.536 & 0.433 & 0.334& 2.905 \\
      LLaVA-Video-7B  & 0.876 & 0.701 & 0.542 & 0.438 & 0.340& 2.898 \\
      \textbf{\model-7B } & \textbf{0.931} & \textbf{0.848} &\textbf{0.768} & \textbf{0.703} & \textbf{0.634}& \textbf{3.885} \\
    \bottomrule
    \end{tabular}
\caption{\textbf{Comparison on CALVIN ABC-D.} Entries marked with * are expert VLAs modified in VLM4VLA settings.}
\label{tab:calvin}
% \vspace{-1em}
\end{table}

\begin{table}[t]
\setlength{\tabcolsep}{3.8mm}

\centering
\small
% \begin{tabular}{r r | r r r r | r}
\begin{tabular}{l | c c c c | c}
% {p{3.2cm} p{0.9cm} p{4.2cm} r r r r r r}
\toprule
\textbf{Model (VLM Backbone)} & \textbf{Carrot} & \textbf{Eggplant} & \textbf{Spoon} & \textbf{Cube} & \textbf{Simpler$\uparrow$ } \\
\midrule
  \skyblue\multicolumn{6}{c}{{Expert Vision-Language-Action Models}}\\
\midrule
OpenVLA* (Llama-2)~\cite{openvla}& 4.2 & 0.0 & 0.0 & 12.5 & 4.2\\
pi0* (Paligemma-1)~\cite{pi0}& 62.5 & 100.0 & 54.2 & 25.0 & 60.4  \\
ThinkAct* (Qwen2.5VL-7B)~\cite{thinkact}& 37.5 & 70.8 & 58.3 & 8.7 & 43.8 \\
\midrule
  \skyblue\multicolumn{6}{c}{{VLM with VLM4VLA Models (Full Fine-tuing)}}\\
\midrule
% Qwen2.5VL-3B&3.8B& 20.8 & 91.7 & 79.2 & 0.0 & 48.0  \\
Qwen2.5VL-7B~\cite{qwen25vl} & 12.5 & 100.0 & 75.0 & 0.0 & 46.8  \\
% Qwen3VL-2B &2.1B&20.8&95.8&79.2&0.0&49.0\\
% Qwen3VL-4B &4.4B&54.2&95.8&75.0&0.0&56.3\\
Qwen3VL-8B~\cite{qwen3vl} &58.3&95.8&79.2&0.0&58.3\\
% Qwen3VL-30B-A3B &31.1B&29.2&79.2&70.8&0.0&44.8\\
Paligemma-1-3B~\cite{paligemma}& 50.0 & 91.7 & 75.0 & 4.2 & 55.3 \\
Paligemma-2-3B~\cite{paligemma2}& 75.0 & 75.0 & 79.2 & 0.0 & 57.3 \\
KosMos-2-1.7B~\cite{Kosmos2}& 37.5 & 100.0 & 75.0 & 29.2 & 60.4 \\
% InternVL3.5-4B & 4.7B& 12.5 & 100.0 & 62.5 & 54.2 & 57.3 & 62.8 \\
\midrule
  \skyblue\multicolumn{6}{c}{{VLM with VLM4VLA Models (Frozen Vision)}}\\
\midrule
Qwen2.5VL-7B  & 4.2 & 66.6 & 4.2 & 0.0 & 18.5 \\
LLaVA-Video-7B  & 0.0 & \textbf{87.5} & 33.3 & 0.0 & 30.2 \\
\textbf{\model-7B} & \textbf{33.3} & 83.3 & \textbf{41.7} & \textbf{25.0} & \textbf{45.8} \\
\bottomrule
\end{tabular}
\caption{\textbf{Comparison on SimplerEnv-Bridge.} Entries marked with * are expert VLAs modified in VLM4VLA settings.}
\label{tab:bridge}
\end{table}

\subsection{Ablation Study}

To investigate the effects of multitask learning, we conduct experiments with different task configurations. For computational efficiency, all models in this study are evaluated using the checkpoints of Stage-1 pre-training {at $224 \times 224$ resolution}. 
Tab.~\ref{tab:ablation} demonstrates the indispensable role of each pre-training objective: 
\textbf{(i) video modeling (\textit{w/o VideoSSL}):} 
Omitting video data from $\mathcal{L}_{\text{ssl}}$ severely degrades dynamic perception ({SSv2 drops from $69.3\%$ to $63.0\%$; DAVIS declines from $71.6$ to $67.7$) and embodied control (CALVIN drops from 3.80 to 3.42), confirming its necessity for capturing dynamic motions. 
\textbf{(ii) geometric reconstruction (\textit{w/o 3D Geometry}):} Disabling the streaming geometric reconstruction objective collapses spatial perception (NYUv2 RMSE worsens to 0.471; ADE20K drops $7.3$ mIoU). Crucially, this geometric deprivation causes a sharp decline in downstream spatial intelligence (VSI-Bench drops $4.8\%$) and VLA performance (CALVIN drops $0.46$), validating that explicit 3D priors are prerequisites for Embodied AI. 
\textbf{(iii) vision-language alignment (\textit{w/o Captioning}):} While pure vision probing remains stable, omitting the captioning task causes catastrophic failures in VLM integration (VideoMME and VSI-Bench plummet by $9.1\%$ and $12.4\%$, respectively) and further impacts VLA policy performance. This highlights that early vision-language alignment is critical to bridging the semantic gap.

In conclusion, our unified multitask formulation is not merely a concatenation of independent losses, but a synergistic framework. The semantic (Captioning), dynamic (VideoSSL), and geometric (3D Reconstruction) objectives mutually reinforce the backbone, culminating in a robust representation that excels across both perceptual and embodied tasks.

\begin{table}[t]
    \centering
    \small
    
    \resizebox{\textwidth}{!}{
    \begin{tabular}{l | c c | c c c | c c c}
    \toprule
             &\multicolumn{2}{c|}{\bf\it Video Probing} & \multicolumn{3}{c|}{\bf\it Image Probing} & \multicolumn{3}{c}{\bf\it VLM \& VLA Probing} \\
             \bf Method & \bf SSv2  & \bf DAVIS'17 & \bf ImageNet & \bf NYUv2 & \bf ADE20k & \bf VSIBench & \bf VideoMME & \bf CALVIN \\
            &ACC@1$\uparrow$ & {$\mathcal J\&F$}-Mean$\uparrow$ & ACC@1$\uparrow$ & RMSE$\downarrow$ & mIoU$\uparrow$ & Acc$\uparrow$ & Acc$\uparrow$ & Seq Len$\uparrow$ \\
            \midrule
Ours& \textbf{69.3} & \textbf{71.6} & \underline{85.2} & \textbf{0.379} & \textbf{49.6} & \underline{57.3} & \underline{54.1} & \textbf{3.80} \\ 
w/o. VideoSSL & 63.0 & 67.7 & \textbf{85.4}&0.420 & \underline{47.2} & \textbf{57.9 }& \textbf{55.8} & \underline{3.42}  \\
w/o. 3D Geometry & \underline{68.4} & 69.7 & 85.0 & 0.471 & 42.3 & 52.5 & 53.8 & 3.34 \\
w/o. Captioning &  67.4& \underline{71.0} & 84.4 & \underline{0.395} & 46.9 & 44.9 & 45.0 & 2.38 \\
\bottomrule
\end{tabular}
}
\caption{{\bf Ablation study for pre-training tasks.} We investigate the contribution of each objective during pre-training stage: \textit{w/o VideoSSL} removes video data in $\mathcal{L}_{ssl}$, \textit{w/o 3D Geometry} omits streaming geometric reconstruction, and \textit{w/o Captioning} disables vision-language alignment.} 
\label{tab:ablation}
% \vspace{-1em}
\end{table}

\section{Conclusion}
In this paper, we introduce \model, a unified vision foundation model that integrates causal spatiotemporal attention and 3D-RoPE into a pre-trained Vision Transformer. Through multi-task training on large-scale datasets from diverse sources, our model achieves competitive performance across a broad spectrum of tasks, including image and video probing, streaming geometric reconstruction, general and spatial video question answering, and robotic manipulation. While \model does not uniformly surpass specialized state-of-the-art methods on every benchmark, its consistent competitiveness across such diverse tasks underscores the viability of training a single, versatile vision backbone that generalizes across semantic, spatial, and temporal reasoning. We believe this versatility represents a more meaningful step toward general-purpose visual understanding than benchmark-specific dominance. As our study focuses on validating this unified paradigm, we leave model scaling as a promising future direction that we expect will further close the remaining gaps with task-specific approaches. We hope this work establishes a solid foundation for streaming visual perception, paving the way for more capable interactive and embodied agents.

\clearpage
\begin{center}
    {\large \bfseries \model: Mastering Perception,\\ Reconstruction and Action in Continuous Streams\\[10pt]}
    {\large Appendix}
\end{center}
\renewcommand{\thesection}{\Alph{section}}
\setcounter{section}{0}

\section{Details of the SSL Objective}
We formally give the definitions of our SSL objective, which is a combination of \textbf{DINO loss}~\cite{dino}, \textbf{iBOT loss~\cite{ibot}}, \textbf{KoLeo regularizer}~\cite{koleo}, and \textbf{Gram anchoring}~\cite{dinov3}.
\begin{itemize}
    \item \textbf{DINO loss.} To capture the global semantic information of the video clip (or image), we first aggregate the sequence of class tokens $\{ \mathbf{z}_{\text{cls}, t} \}_{t=1}^T$ via temporal mean pooling: $\bar{\mathbf{z}}_{\text{cls}} = \frac{1}{T} \sum_{t=1}^T \mathbf{z}_{\text{cls}, t}$.
    This aggregated representation is then projected to ``prototype scores'' by the DINO head of the student and teacher, respectively. After applying softmax to get $p_s$ and softmax with Sinkhorn-Knopp centering for $p_t$, the DINO loss minimizes the cross-entropy between these distributions:
    \begin{equation}
        \mathcal{L}_{\text{DINO}} = - \sum_{} p_t \log p_s.
    \end{equation}
    \item \textbf{iBOT loss.} We mask both image and video local patches given to the student, where different masking strategies are applied accordingly. For image patches, we follow DINO's protocol and apply random masks, while the video patches are masked using \textit{multi-block protocol}~\cite{vjepa}. We then apply the iBOT head to the student's masked patches and the teacher's corresponding to the ones masked in the student. Next, we apply softmax and centering as in the \textbf{DINO Loss}, and the iBOT objective is: 
    \begin{equation}
        \mathcal{L}_{\text{iBOT}} = - \sum_{i}^{N_{mask}} p_{ti} \log p_{si},
    \end{equation}
    where $i$ are the indices for masked tokens, and $N_{mask}$ is the total number of masks in the sample.
    \item \textbf{KoLeo regularizer.} The KoLeo regularizer encourages a uniform distribution of features within a batch. Given a batch of vectors $\left(x_1, ..., x_n \right)$, the KoLeo loss is defined as:
    \begin{equation}
        \mathcal{L}_{\text{KoLeo}} = -\frac{1}{n} \sum_{i=1}^{n} \log(\rho_{n, i}),
    \end{equation}
    where $\rho_{n,i} = \min_{j \neq i} \|x_i - x_j \|$ is the minimum distance of $x_i$ to any other point within the batch.
    \item \textbf{Gram anchoring.} To maintain a consistent patch-level feature during training, we introduce a loss that operates on the gram matrix. The gram matrix is defined as the matrix of all pairwise dot products within the current image. At the beginning of the training, we select the pre-trained DINOv3 as our gram teacher. Given an image sequence of $hw$ patches, $\mathbf{X}_s\in \mathbb{R}^{hw\times d}$ denotes the L2-normalized local patch features, and correspondingly for the gram teacher's local patch features $\mathbf{X}_g\in \mathbb{R}^{hw\times d}$. We define the loss as:
    \begin{equation}
        \mathcal{L}_{\text{gram}} = \| \mathbf{X}_s \cdot \mathbf{X}_s^\top - \mathbf{X}_g \cdot \mathbf{X}_g^\top \|_{\text{F}}^2.
    \end{equation}
    Note that we only enable this term for the global crops of the image inputs. 
\end{itemize}
As mentioned in the main text, we optimize the following loss for \textbf{static and temporal representation learning}:
\begin{equation}
    \mathcal{L}_{\mathrm{ssl}} =
    \mathcal{L}_{\mathrm{DINO}} +
    \mathcal{L}_{\mathrm{iBOT}} +
    0.1 \times \mathcal{L}_{\mathrm{KoLeo}} + \mathcal{L}_{\mathrm{gram}},
\end{equation}

\section{Computational Efficiency and Scaling Analysis}
\label{sec:efficiency}

To demonstrate the real-world deployment viability and the efficiency of our causal streaming design, we conduct a comprehensive computational analysis comparing \model (utilizing the causal KV-cache) against a standard bi-directional recomputation baseline. The evaluation is performed on a single NVIDIA H800 GPU with a batch size of 1 and an input resolution of $224 \times 224$. Our attention mechanism is implemented using Scaled Dot Product Attention (SDPA) leveraging FlashAttention~\cite{flashattn} kernels. 

\vspace{3pt} \noindent \textbf{Temporal Context Expansion and Memory Efficiency.}
The ability to process video streams under tight resource constraints is a core strength of \model. We benchmark the per-frame inference latency and peak VRAM consumption for the $T$-th frame given a historical context of $T-1$ frames (Fig.~\ref{fig:kvcache} and Tab.~\ref{tab:inference_efficiency}).
% (Tab.~\ref{tab:inference_efficiency}). 

For the baseline employing full bi-directional recomputation, both latency and memory usage grow prohibitively. Specifically, the recomputation latency reaches nearly 1 second (0.998s) at $T=64$, and the system encounters a catastrophic Out-of-Memory (OOM) error at $T=128$ due to the quadratic growth of the attention map. 

In contrast, \model utilizes a persistent KV-cache mechanism, resulting in a significantly optimized resource profile. At $T=64$, our model is approximately 15$\times$ faster than the baseline with only 0.067s latency. More importantly, \model maintains a linear and manageable memory growth, effectively preventing OOM even when the context extends to $T=512$. This efficiency ensures that the backbone leaves sufficient VRAM for downstream VLM heads or real-time policy execution on embodied platforms.

\begin{table}[ht]
    \centering
    \small 
    
    \setlength{\tabcolsep}{2.2mm} 
    \begin{tabular}{ll | c c c c c c}
        \toprule
        \multicolumn{2}{l|}{\textbf{Context Length ~$T$}} & \textbf{16} & \textbf{32} & \textbf{64} & \textbf{128} & \textbf{256} & \textbf{512}\\
        \midrule
        Full Recompute & Latency (s) & 0.125 & 0.329 & 0.998 & N/A & N/A &  N/A \\
        (Bi-directional) & Memory (GB) & 3.46 & 5.95 & 15.14 & OOM & OOM &  OOM \\
        \midrule
        \rowcolor[RGB]{250, 235, 215} \textbf{\model} & \textbf{Latency (s)} & \textbf{0.042} & \textbf{0.057} & \textbf{0.067} & \textbf{0.115} & \textbf{0.216} & \textbf{0.414} \\
        \rowcolor[RGB]{250, 235, 215} \textbf{(KV-cache)} & \textbf{Memory (GB)} & {3.60} & \textbf{4.81} & \textbf{7.26} & \textbf{12.14} & \textbf{21.92} & \textbf{41.46} \\
        \bottomrule
    \end{tabular}
    \caption{\textbf{Inference Efficiency with Varying Context Lengths.} We report the per-frame latency (s) and peak VRAM (GB) for the $T$-th frame. Bi-di recomputes the entire sequence, while \model uses a persistent KV-cache.}
    \label{tab:inference_efficiency}
\end{table}

\begin{figure}[h]
    \centering
    \includegraphics[width=1.0\textwidth]{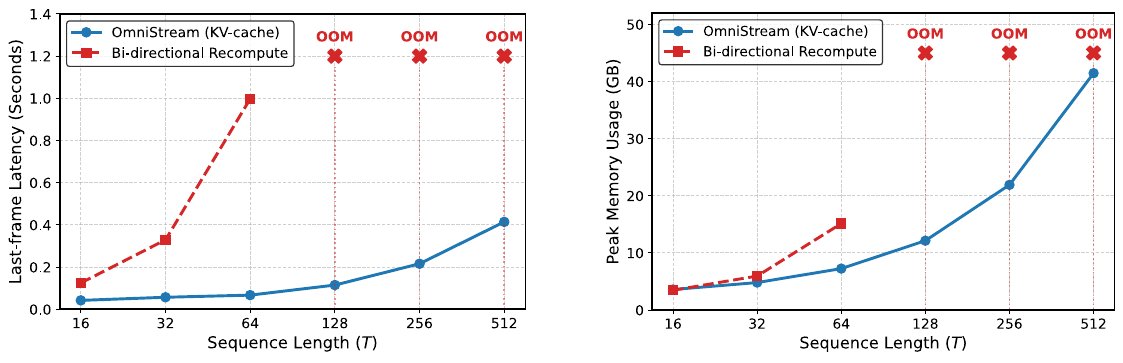}
    \caption{\textbf{Qualitative results on Sintel video depth reconstruction.} Our model maintains temporal coherence across long sequences.}
    \label{fig:kvcache}
\end{figure}

\section{Details of Downstream Tasks}
\subsection{Image Probing Details}
To evaluate the static representation quality of our model, we perform linear probing on three standard benchmarks: ImageNet-1K~\cite{imagenet} for image classification, NYUv2~\cite{NYUv2} for monocular depth estimation, and ADE20K~\cite{ade20k} for semantic segmentation. For all tasks, we train a task-specific linear head on top of the strictly frozen visual backbone.

We closely follow the evaluation protocols and configurations of DINOv3~\cite{dinov3}. Detailed hyperparameters are summarized in Tab.~\ref{tab:linear_probe_results}. Note that for ImageNet classification, we conduct a hyperparameter sweep over 13 learning rate configurations, ranging from $1 \times 10^{-5}$ to $1 \times 10^{-1}$, to report the optimal performance.

\begin{table}[htbp]
    \centering
    \small
    \setlength{\tabcolsep}{1.5mm}
    \begin{tabular}{lccc}
        \toprule
        & \multicolumn{3}{c}{\textbf{Dataset Name}} \\
        \cmidrule(lr){2-4}
        \textbf{Hyperparameters} & \textbf{ImageNet-1K} & \textbf{NYUv2} & \textbf{ADE-20K} \\
        \midrule
        Resolution     & $224 \times 224$ & $416 \times 544$ & $512 \times 512$ \\
        Learning Rate  & $ \{ 1 \times 10^{-5},...,1 \times 10^{-1} \} $& $1 \times 10^{-3}$ & $1 \times 10^{-3}$ \\
        Warm-up Steps  & $0$           & $12,800$           & $1,500$           \\
        Total Steps    & $12,500$         & $38,400$          & $40,000$          \\
        \bottomrule
    \end{tabular}
    \caption{Linear probing detailed configurations.}
    \label{tab:linear_probe_results}
\end{table}

\subsection{Video Probing Details}
To evaluate the dynamic spatiotemporal representation quality of our model, we perform attentive probing on two standard video action recognition benchmarks: Kinetics-400 (K400)~\cite{k400} and Something-Something V2 (SSv2)~\cite{ssv2}. Following V-JEPA 2~\cite{vjepa2}, we train a lightweight attentive probe—consisting of 4 Transformer blocks with 16 attention heads—on top of the strictly frozen visual backbone.

During training, we sample 16 frames with a frame stride of 4 for both datasets. During inference, we follow the standard multi-view testing protocol, taking multiple temporal segments and spatial crops to obtain the final video-level prediction. Detailed configurations and evaluation protocols are summarized in Tab.~\ref{tab:video_probe_results}.
\begin{table}[htbp]
    \centering
    \small
    \setlength{\tabcolsep}{1mm}
    \begin{tabular}{lcc}
        \toprule
        & \multicolumn{2}{c}{\textbf{Dataset Name}} \\
        \cmidrule(lr){2-3}
        \textbf{Hyperparameters} & \textbf{Kinetics-400} & \textbf{Something-Something V2} \\
        \midrule
        Input Resolution        & $256 \times 256$ & $256 \times 256$ \\
        Frames $\times$ Stride  & $16 \times 4$    & $16 \times 4$    \\
        Training Epochs         & $20$             & $20$             \\
        Warm-up Epochs          & $0$              & $0$              \\
        Learning Rate     & $\{1\times 10^{-4}, 3\times 10^{-4}\}$  & $3\times 10^{-4}$  \\
        \midrule
        Test Segments $\times$ Crops & $8 \times 3$ & $2 \times 3$ \\
        \bottomrule
    \end{tabular}
    \caption{Attentive probing detailed configurations for video action recognition.}
    \label{tab:video_probe_results}
\end{table}

Furthermore, we evaluate dense spatiotemporal tracking on the DAVIS'17~\cite{davis17} Video Object Segmentation (VOS) benchmark, utilizing the standard 480p resolution. 

During the dense feature extraction stage, we employ distinct inference strategies tailored to each model's architectural characteristics: For \model, we fully leverage our causal KV-cache mechanism to encode the entire video sequence in a single continuous stream. This allows the model to maintain long-term temporal coherence effortlessly. In contrast, for the V-JEPA 2 baseline, which relies on a rigid spatiotemporal tubelet size of $2 \times 16 \times 16$, we conduct the evaluation using a 16-frame sliding window. To maintain frame-by-frame temporal consistency and accommodate its temporal downsampling mechanism, we explicitly duplicate the input frames. Lastly, for the DINOv3 baseline, we simply adopt a standard per-frame feature extraction.

\subsection{VLM Probing Details}

\begin{table}[ht]
\centering
\small
\resizebox{\textwidth}{!}{
\renewcommand{\arraystretch}{1.2}
    \begin{tabular}{cl|c|c|c|c}
    \toprule
    & & \multirow{2}{*}{\textbf{Stage-1}} & \multirow{2}{*}{\textbf{Stage-1.5}} & \multicolumn{1}{c|}{\textbf{Stage-2}} & \multicolumn{1}{c}{\textbf{Stage-3}}\\
    \cmidrule(lr){5-6}
    & & & & \textbf{Single-Image} & \textbf{Video} \\
    \midrule
    \multirow{2}{*}{\rotatebox{90}{Vision}} & \textbf{Resolution} & 512 & $512 \times \{2\times2, 1\times\{2,3\}, \{2,3\}\times1\}$ & $512 \times \{\{1\times1\}, \cdots, \{6\times6\}\}$ & $512 \times 32~(Frames)$ \\
    & \textbf{\#Tokens} & 1024 & Max $1024 \times 5$ & Max $1024 \times 10$ & Max $1024/4 \times 32$  \\
    \midrule
    \multirow{2}{*}{\rotatebox{90}{Data}} & \textbf{Dataset} & LCS & LLaVA-OV-Caption~\cite{llava_one_vision} & LLaVA-OV-SI~\cite{llava_one_vision} & LLaVA-Video~\cite{llavavideo} \& Spatial QA \\
    & \textbf{\#Samples} & 558K & 4M & 3.2M & 2.9M + 1.1M \\
    \midrule
    \multirow{2}{*}{\rotatebox{90}{Model}} & \textbf{Trainable} & Projector & Projector + LLM & Projector + LLM & Projector + LLM \\
    & Qwen2.5-7B & 20.0M & 7.6B & 7.6B & 7.6B \\
    \midrule
    \multirow{4}{*}{\rotatebox{90}{Training}} & \textbf{Batch Size} & 512 & 256 & 256 & 128 \\
    & \textbf{LR: $\psi_{\text{vision}}$} & N/A &  N/A &  N/A &  N/A \\
    & \textbf{LR: $\{\theta_{\text{proj}}, \phi_{\text{LLM}}\}$} & $1 \times 10^{-3}$ & $1 \times 10^{-5}$ & $1 \times 10^{-5}$ & $1 \times 10^{-5}$ \\
    & \textbf{Epoch} & 1 & 1 & 1 & 1 \\
    \bottomrule
    \end{tabular}
    }
    \caption{Detailed configuration for each training stage of our \model-7B.}
    \label{tab:vlm_config}
\end{table}
During the VLM training experiments, we adopt the LLaVA-Video~\cite{llavavideo} pipeline and freeze our vision model throughout all training stages, as given in Tab.~\ref{tab:vlm_config}. We first perform a three-stage fine-tuning following LLaVA-SI~\cite{llava_one_vision}, which is also the base model for LLaVA-Video. We make sure the same exposure to general video data as LLaVA-Video, where we use a mixture of datasets of LLaVA-Video-178K~\cite{llavavideo}, ActivityNet-QA~\cite{anetqa}, NExT-QA~\cite{nextqa}, Perception Test~\cite{perceptiontest}, LLaVA-Hound-255K~\cite{llavahound}, and LLaVA-OV-1.1M~\cite{llava_one_vision}. For spatial video QA, we use a combination of VLM-3R-200K~\cite{vlm3r}, VSI-590K~\cite{cambrian-s}, and ViCA-322K~\cite{vica}, resulting in a total of 1.1M spatial video QA samples.

We set the base resolution of our visual encoder to $512$, which is how our backbone is trained, and adopt the AnyRes-Max strategy~\cite{llava_one_vision} for image inputs. For videos, we apply a $2\times2$ average pooling function to each frame and sample at FPS=1 with max frames during training being 32. For VLM training in our ablation study, all stages except stage 1 are sub-sampled at 10\% of the total training samples. We run all experiments using 32 $\times$ NVIDIA H200 GPUs.

\subsection{VLA Probing Details}
\begin{table}[h]
    \centering
    \small
    \setlength{\tabcolsep}{4mm}
    \begin{tabular}{lcc}
        \toprule
        & \multicolumn{2}{c}{\textbf{Benchmark}} \\
        \cmidrule(lr){2-3}
        \textbf{Hyperparameters} & \textbf{CALVIN (ABC-D)} & \textbf{BridgeV2} \\
        \midrule
        Input Resolution        & $224 \times 224$ & $224 \times 224$ \\
        Action Space            & Continuous (7-DoF) & Continuous (7-DoF) \\
        \midrule
        Optimizer               & Adam             & Adam \\
        Learning Rate           & $2 \times 10^{-5}$ & $2 \times 10^{-5}$ \\
        Projector Learning Rate & $1 \times 10^{-4}$ & $1 \times 10^{-4}$ \\
        Gradient Clip Value     & $1.0$            & $1.0$ \\
        Batch Size per GPU      & 8                & 8 \\
        Gradient Accumulation   & 2                & 8 \\
        Total Epochs            & $5$              & $5$ \\
        \bottomrule
    \end{tabular}
    \caption{Detailed VLA fine-tuning configurations for CALVIN and BridgeV2.}
    \label{tab:vla_config}
\end{table}
To evaluate the capability of our model in embodied control, we adopt the training pipeline introduced by VLM4VLA~\cite{vlm4vla}. Specifically, we evolve our integrated Vision-Language Model (\model-7B, built upon Qwen2.5-7B-Instruct) into a Vision-Language-Action (VLA) model. This is achieved by appending a learnable query token at the end of the input sequence to aggregate action-oriented semantics, followed by a lightweight MLP action head to decode continuous 7-DoF actions.

In our \textit{frozen vision} experiments, we strictly freeze the visual backbone to evaluate the quality of its pre-trained representations. During the VLA fine-tuning phase, we optimize the multimodal projector, the Large Language Model (LLM), the text embeddings, and the newly initialized action head. 

We conduct evaluations on two standard robotic manipulation benchmarks: CALVIN (Task ABC-D) and SimplerEnv-BridgeV2. The training is performed using the Adam optimizer with bfloat16 precision and DeepSpeed ZeRO-2 optimization. The detailed hyperparameters for both benchmarks are summarized in Tab.~\ref{tab:vla_config}. We conduct all VLA experiments using 8 $\times$ NVIDIA H200 GPUs.

\section{More Qualitative Results}
\subsection{Video Depth Visualizations}
Fig.~\ref{fig:vdepth_vis} presents qualitative visualizations of streaming video depth prediction on Sintel\cite{sintel}. As shown, \model effortlessly maintains high-fidelity spatial details and long-term temporal consistency across highly dynamic scenes.
\begin{figure}[h]
    \centering
    \includegraphics[width=1.0\textwidth]{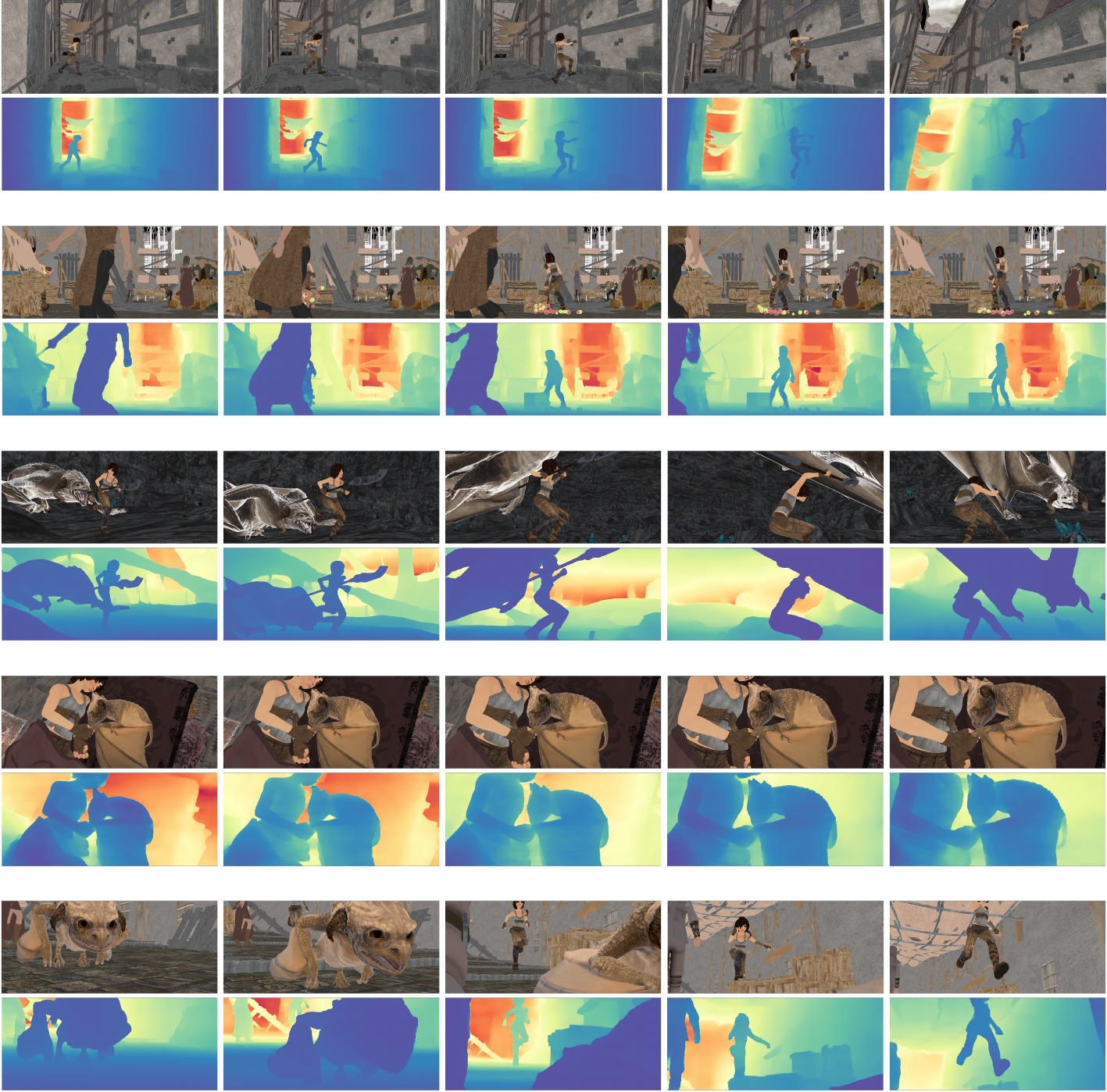}
    \vspace{-0.6cm}
    \caption{\textbf{Qualitative results on the Sintel video depth.}}
    \label{fig:vdepth_vis}
    \vspace{-0.6cm}
\end{figure}

\subsection{DAVIS'17 VOS Visualizations}
Qualitative results for mask propagation on DAVIS'17~\cite{davis17} are illustrated in Fig.~\ref{fig:davis_vos}. By leveraging our causal KV-cache for continuous encoding, \model exhibits a pronounced advantage over V-JEPA 2. It successfully tracks intricate object shapes through fast appearance changes, whereas V-JEPA 2 frequently loses dense spatial alignment under sliding-window inference.

\begin{figure}[htbp]
    \centering
    \begin{subfigure}{\textwidth}
        \includegraphics[width=\textwidth]{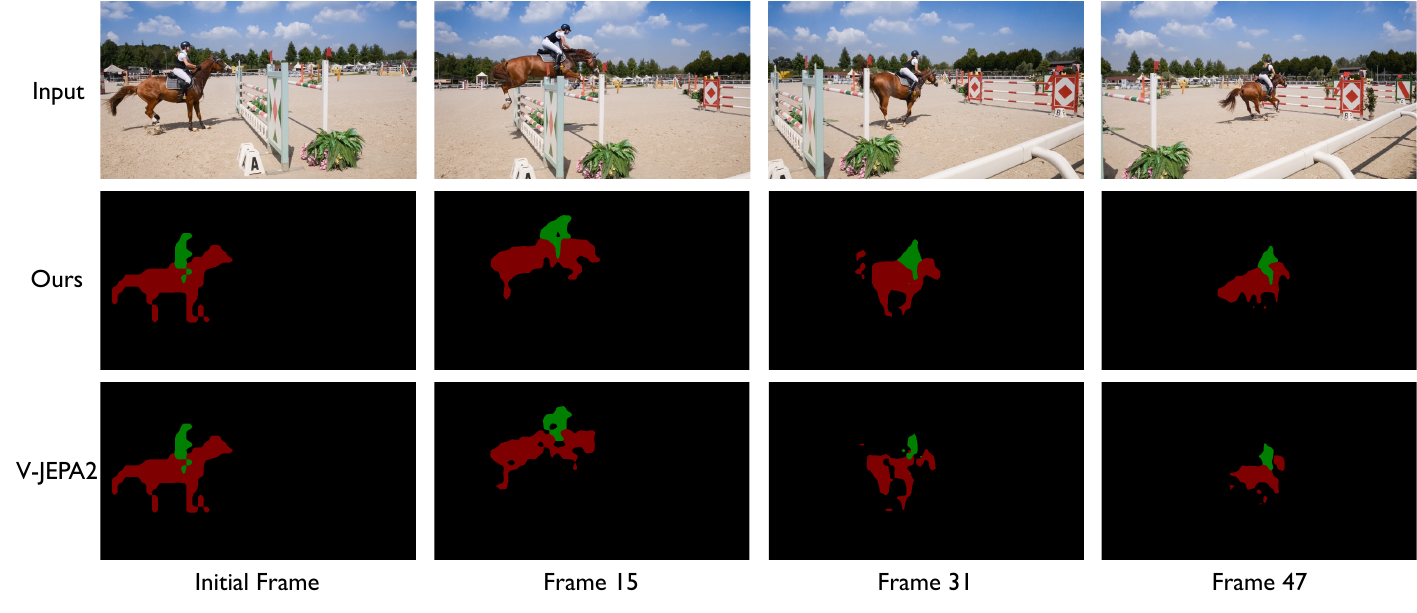}
        \vspace{-0.5cm}
        \caption{{horsejump-high sequence.}}
        \label{fig:davis_horse}
    \end{subfigure}
    
    \vspace{0.2cm}
    \begin{subfigure}{\textwidth}
        \includegraphics[width=\textwidth]{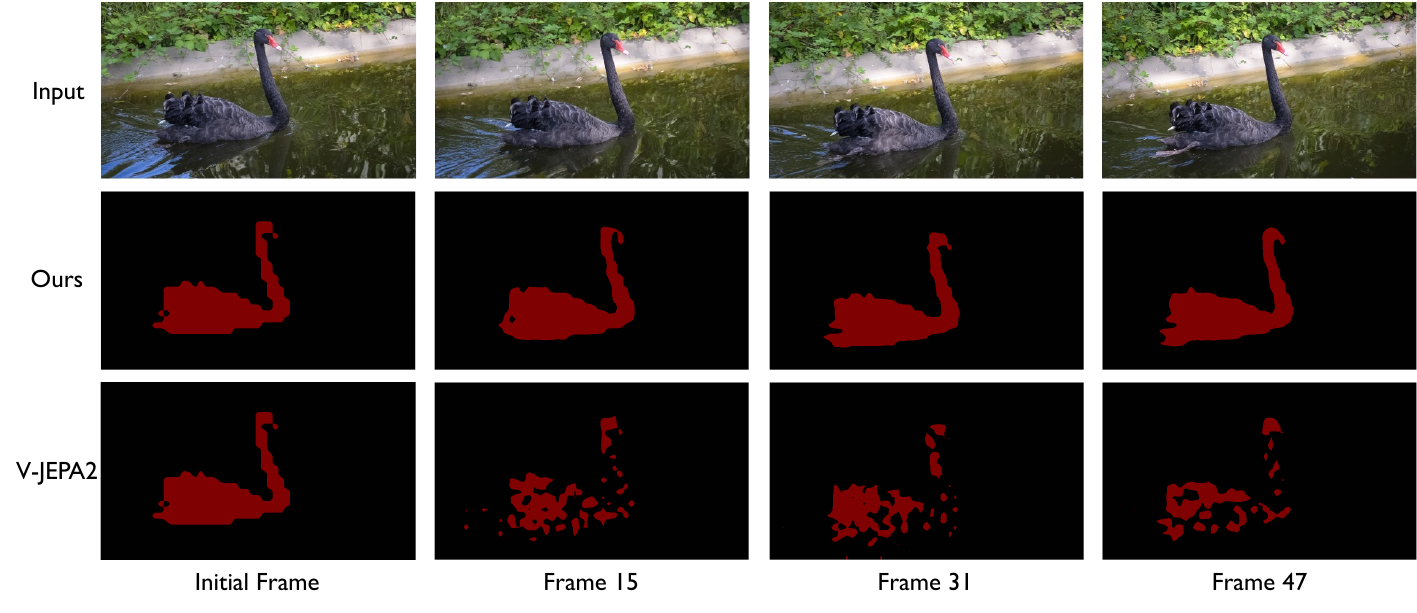}
        \vspace{-0.5cm}
        \caption{{blackswan sequence.}}
        \label{fig:davis_swan}
    \end{subfigure}
    
    \vspace{0.2cm}
    \begin{subfigure}{\textwidth}
        \includegraphics[width=\textwidth]{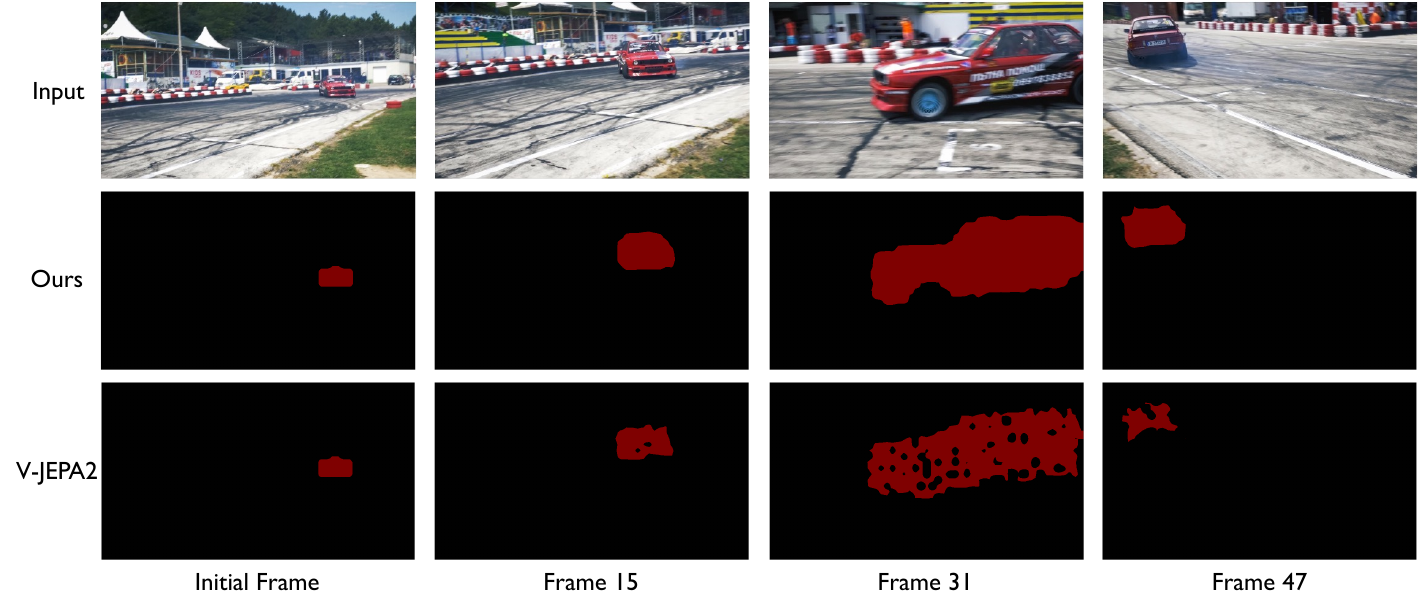}
        \vspace{-0.5cm}
        \caption{{drift-straight sequence.}}
        \label{fig:davis_drift}
    \end{subfigure}
    
    \vspace{-0.2cm}
    \caption{
        \textbf{Qualitative results on the DAVIS'17 benchmark.} 
    }
    \label{fig:davis_vos}
    \vspace{-0.6cm}
\end{figure}

\clearpage

{
    \small
    \bibliographystyle{ieeenat_fullname}
    \bibliography{reference}

@String(CVPR= {IEEE Conf. Comput. Vis. Pattern Recog.})

@String(ICCV= {Int. Conf. Comput. Vis.})

@String(ECCV= {Eur. Conf. Comput. Vis.})

@String(ICLR = {Int. Conf. Learn. Represent.})

@String(AAAI = {AAAI})

@String(CVPR  = {CVPR})

@String(ICCV  = {ICCV})

@String(ECCV  = {ECCV})

@String(ICLR  = {ICLR})

@article{datacomp,
  title={Datacomp: In search of the next generation of multimodal datasets},
  author={Gadre, Samir Yitzhak and Ilharco, Gabriel and Fang, Alex and Hayase, Jonathan and Smyrnis, Georgios and Nguyen, Thao and Marten, Ryan and Wortsman, Mitchell and Ghosh, Dhruba and Zhang, Jieyu and others},
  journal={NeurIPS},
  year={2023}
}

@article{imagenet21k,
  title={Imagenet-21k pretraining for the masses},
  author={Ridnik, Tal and Ben-Baruch, Emanuel and Noy, Asaf and Zelnik-Manor, Lihi},
  journal={arXiv preprint arXiv:2104.10972},
  year={2021}
}

@inproceedings{ssv2,
  title={The" something something" video database for learning and evaluating visual common sense},
  author={Goyal, Raghav and Ebrahimi Kahou, Samira and Michalski, Vincent and Materzynska, Joanna and Westphal, Susanne and Kim, Heuna and Haenel, Valentin and Fruend, Ingo and Yianilos, Peter and Mueller-Freitag, Moritz and others},
  booktitle={ICCV},
  year={2017}
}

@article{PerceptionEncoder,
  title={Perception encoder: The best visual embeddings are not at the output of the network},
  author={Bolya, Daniel and Huang, Po-Yao and Sun, Peize and Cho, Jang Hyun and Madotto, Andrea and Wei, Chen and Ma, Tengyu and Zhi, Jiale and Rajasegaran, Jathushan and Rasheed, Hanoona and others},
  journal={arXiv preprint arXiv:2504.13181},
  year={2025}
}

@article{Kosmos2,
  title={Kosmos-2: Grounding Multimodal Large Language Models to the World},
  author={Zhiliang Peng and Wenhui Wang and Li Dong and Yaru Hao and Shaohan Huang and Shuming Ma and Furu Wei},
  journal={ArXiv},
  year={2023},
}

@misc{coyo-700m,
  title         = {COYO-700M: Image-Text Pair Dataset},
  author        = {Byeon, Minwoo and Park, Beomhee and Kim, Haecheon and Lee, Sungjun and Baek, Woonhyuk and Kim, Saehoon},
  year          = {2022},
  howpublished  = {\url{https://github.com/kakaobrain/coyo-dataset}}
}

@inproceedings{refcoco,
  title={Referitgame: Referring to objects in photographs of natural scenes},
  author={Kazemzadeh, Sahar and Ordonez, Vicente and Matten, Mark and Berg, Tamara},
  booktitle={EMNLP},
  year={2014}
}

@article{blip3,
  title={xgen-mm (blip-3): A family of open large multimodal models},
  author={Xue, Le and Shu, Manli and Awadalla, Anas and Wang, Jun and Yan, An and Purushwalkam, Senthil and Zhou, Honglu and Prabhu, Viraj and Dai, Yutong and Ryoo, Michael S and others},
  journal={arXiv preprint arXiv:2408.08872},
  year={2024}
}

@article{chen2024far,
  title={How Far Are We to GPT-4V? Closing the Gap to Commercial Multimodal Models with Open-Source Suites},
  author={Chen, Zhe and Wang, Weiyun and Tian, Hao and Ye, Shenglong and Gao, Zhangwei and Cui, Erfei and Tong, Wenwen and Hu, Kongzhi and Luo, Jiapeng and Ma, Zheng and others},
  journal={arXiv preprint arXiv:2404.16821},
  year={2024}
}

@inproceedings{co3d,
  title={Common objects in 3d: Large-scale learning and evaluation of real-life 3d category reconstruction},
  author={Reizenstein, Jeremy and Shapovalov, Roman and Henzler, Philipp and Sbordone, Luca and Labatut, Patrick and Novotny, David},
  booktitle={ICCV},
  year={2021}
}

@inproceedings{wildrgbd,
  title={Rgbd objects in the wild: Scaling real-world 3d object learning from rgb-d videos},
  author={Xia, Hongchi and Fu, Yang and Liu, Sifei and Wang, Xiaolong},
  booktitle={CVPR},
  year={2024}
}

@article{blendedmvs,
  title={BlendedMVS: A Large-scale Dataset for Generalized Multi-view Stereo Networks},
  author={Yao, Yao and Luo, Zixin and Li, Shiwei and Zhang, Jingyang and Ren, Yufan and Zhou, Lei and Fang, Tian and Quan, Long},
  journal={CVPR},
  year={2020}
}

@inproceedings{megadepth,
  	title={MegaDepth: Learning Single-View Depth Prediction from Internet Photos},
  	author={Zhengqi Li and Noah Snavely},
  	booktitle={CVPR},
  	year={2018}
}

@inproceedings{mvssynth,
  author       = "Huang, Po-Han and Matzen, Kevin and Kopf, Johannes and Ahuja, Narendra and Huang, Jia-Bin",
  title        = "DeepMVS: Learning Multi-View Stereopsis",
  booktitle    = "CVPR",
  year         = "2018"
}

@inproceedings{pointodyssey,
    author = {Yang Zheng and Adam W. Harley and Bokui Shen and Gordon Wetzstein and Leonidas J. Guibas},
    title = {PointOdyssey: A Large-Scale Synthetic Dataset for Long-Term Point Tracking},
    booktitle = {ICCV},
    year = {2023}
}

@inproceedings{hypersim,
  title={Hypersim: A photorealistic synthetic dataset for holistic indoor scene understanding},
  author={Roberts, Mike and Ramapuram, Jason and Ranjan, Anurag and Kumar, Atulit and Bautista, Miguel Angel and Paczan, Nathan and Webb, Russ and Susskind, Joshua M},
  booktitle={ICCV},
  year={2021}
}

@inproceedings{tartanair,
  title =   {TartanAir: A Dataset to Push the Limits of Visual SLAM},
  author =  {Wang, Wenshan and Zhu, Delong and Wang, Xiangwei and Hu, Yaoyu and Qiu, Yuheng and Wang, Chen and Hu, Yafei and Kapoor, Ashish and Scherer, Sebastian},
  booktitle = {IROS},
  year =    {2020}
}

@inproceedings{dl3dv,
  title={Dl3dv-10k: A large-scale scene dataset for deep learning-based 3d vision},
  author={Ling, Lu and Sheng, Yichen and Tu, Zhi and Zhao, Wentian and Xin, Cheng and Wan, Kun and Yu, Lantao and Guo, Qianyu and Yu, Zixun and Lu, Yawen and others},
  booktitle={CVPR},
  year={2024}
}

@inproceedings{scannet,
    title={ScanNet: Richly-annotated 3D Reconstructions of Indoor Scenes},
    author={Dai, Angela and Chang, Angel X. and Savva, Manolis and Halber, Maciej and Funkhouser, Thomas and Nie{\ss}ner, Matthias},
    booktitle = {CVPR},
    year = {2017}
}

@inproceedings{scannetpp,
  title={Scannet++: A high-fidelity dataset of 3d indoor scenes},
  author={Yeshwanth, Chandan and Liu, Yueh-Cheng and Nie{\ss}ner, Matthias and Dai, Angela},
  booktitle={ICCV},
  year={2023}
}

@inproceedings{arkitscenes,
  title={ARKitScenes: A Diverse Real-World Dataset For 3D Indoor Scene Understanding Using Mobile RGB-D Data},
  author={Baruch, Gilad and Chen, Zhuoyuan and Dehghan, Afshin and Feigin, Yuri and Fu, Peter and Gebauer, Thomas and Kurz, Daniel and Dimry, Tal and Joffe, Brandon and Schwartz, Arik and others},
  booktitle={NeurIPS},
  year={2021}
}

@article{vkitti2,
  title={Virtual kitti 2},
  author={Cabon, Yohann and Murray, Naila and Humenberger, Martin},
  journal={arXiv preprint arXiv:2001.10773},
  year={2020}
}

@inproceedings{mapfree,
      title={Map-free Visual Relocalization: Metric Pose Relative to a Single Image},
      author={Arnold, Eduardo and Wynn, Jamie and Vicente, Sara and Garcia-Hernando, Guillermo and Monszpart, {\'{A}}ron and Prisacariu, Victor Adrian and Turmukhambetov, Daniyar and Brachmann, Eric},
      booktitle={ECCV},
      year={2022},
}

@article{dynamicreplica,
  author    = {Nikita Karaev and Ignacio Rocco and Benjamin Graham and Natalia Neverova and Andrea Vedaldi and Christian Rupprecht},
  title     = {DynamicStereo: Consistent Dynamic Depth from Stereo Videos},
  journal   = {CVPR},
  year      = {2023},
}

@article{omniworld,
      title={OmniWorld: A Multi-Domain and Multi-Modal Dataset for 4D World Modeling}, 
      author={Yang Zhou and Yifan Wang and Jianjun Zhou and Wenzheng Chang and Haoyu Guo and Zizun Li and Kaijing Ma and Xinyue Li and Yating Wang and Haoyi Zhu and Mingyu Liu and Dingning Liu and Jiange Yang and Zhoujie Fu and Junyi Chen and Chunhua Shen and Jiangmiao Pang and Kaipeng Zhang and Tong He},
      journal={arXiv preprint arXiv:2509.12201},
      year={2025}
}

@InProceedings{spring,
    author    = {Lukas Mehl and Jenny Schmalfuss and Azin Jahedi and Yaroslava Nalivayko and Andr\'es Bruhn},
    title     = {Spring: A High-Resolution High-Detail Dataset and Benchmark for Scene Flow, Optical Flow and Stereo},
    booktitle = {CVPR},
    year      = {2023}
}

@inproceedings{ariasynenv,
  title={Aria digital twin: A new benchmark dataset for egocentric 3d machine perception},
  author={Pan, Xiaqing and Charron, Nicholas and Yang, Yongqian and Peters, Scott and Whelan, Thomas and Kong, Chen and Parkhi, Omkar and Newcombe, Richard and Ren, Yuheng Carl},
  booktitle={ICCV},
  year={2023}
}

@InProceedings{waymo,
    author = {Sun, Pei and Kretzschmar, Henrik and Dotiwalla, Xerxes and Chouard, Aurelien and Patnaik, Vijaysai and Tsui, Paul and Guo, James and Zhou, Yin and Chai, Yuning and Caine, Benjamin and Vasudevan, Vijay and Han, Wei and Ngiam, Jiquan and Zhao, Hang and Timofeev, Aleksei and Ettinger, Scott and Krivokon, Maxim and Gao, Amy and Joshi, Aditya and Zhang, Yu and Shlens, Jonathon and Chen, Zhifeng and Anguelov, Dragomir},
    title = {Scalability in Perception for Autonomous Driving: Waymo Open Dataset},
    booktitle = {CVPR},
    year = {2020}
}

@article{dinov3,
  title={Dinov3},
  author={Sim{\'e}oni, Oriane and Vo, Huy V and Seitzer, Maximilian and Baldassarre, Federico and Oquab, Maxime and Jose, Cijo and Khalidov, Vasil and Szafraniec, Marc and Yi, Seungeun and Ramamonjisoa, Micha{\"e}l and others},
  journal={arXiv preprint arXiv:2508.10104},
  year={2025}
}

@inproceedings{vggt,
  title={VGGT: Visual Geometry Grounded Transformer},
  author={Wang, Jianyuan and Chen, Minghao and Karaev, Nikita and Vedaldi, Andrea and Rupprecht, Christian and Novotny, David},
  booktitle={CVPR},
  year={2025}
}

@inproceedings{cut3r,
    Author = {Qianqian Wang* and Yifei Zhang* and Aleksander Holynski and Alexei A. Efros and Angjoo Kanazawa},
    Title = {Continuous 3D Perception Model with Persistent State},
    Year = {2025},
      booktitle={CVPR},
    }

@article{vjepa,
  title={Revisiting feature prediction for learning visual representations from video},
  author={Bardes, Adrien and Garrido, Quentin and Ponce, Jean and Chen, Xinlei and Rabbat, Michael and LeCun, Yann and Assran, Mahmoud and Ballas, Nicolas},
  journal={arXiv preprint arXiv:2404.08471},
  year={2024}
}

@article{vjepa2,
  title={V-jepa 2: Self-supervised video models enable understanding, prediction and planning},
  author={Assran, Mido and Bardes, Adrien and Fan, David and Garrido, Quentin and Howes, Russell and Muckley, Matthew and Rizvi, Ammar and Roberts, Claire and Sinha, Koustuv and Zholus, Artem and others},
  journal={arXiv preprint arXiv:2506.09985},
  year={2025}
}

@inproceedings{vision_registers,
  title={Vision Transformers Need Registers},
  author={Darcet, Timoth{\'e}e and Oquab, Maxime and Mairal, Julien and Bojanowski, Piotr},
  booktitle={ICLR},
  year={2024}
}

@inproceedings{depthanything1,
      title={Depth Anything: Unleashing the Power of Large-Scale Unlabeled Data}, 
      author={Yang, Lihe and Kang, Bingyi and Huang, Zilong and Xu, Xiaogang and Feng, Jiashi and Zhao, Hengshuang},
      booktitle={CVPR},
      year={2024}
}

@article{depthanything2,
  title={Depth anything v2},
  author={Yang, Lihe and Kang, Bingyi and Huang, Zilong and Zhao, Zhen and Xu, Xiaogang and Feng, Jiashi and Zhao, Hengshuang},
  journal={NeurIPS},
  year={2024}
}

@article{depthanything3,
  title={Depth anything 3: Recovering the visual space from any views},
  author={Lin, Haotong and Chen, Sili and Liew, Junhao and Chen, Donny Y and Li, Zhenyu and Shi, Guang and Feng, Jiashi and Kang, Bingyi},
  journal={arXiv preprint arXiv:2511.10647},
  year={2025}
}

@inproceedings{videomae2,
  title={Videomae v2: Scaling video masked autoencoders with dual masking},
  author={Wang, Limin and Huang, Bingkun and Zhao, Zhiyu and Tong, Zhan and He, Yinan and Wang, Yi and Wang, Yali and Qiao, Yu},
  booktitle={CVPR},
  year={2023}
}

@inproceedings{siglip,
  title={Sigmoid loss for language image pre-training},
  author={Zhai, Xiaohua and Mustafa, Basil and Kolesnikov, Alexander and Beyer, Lucas},
  booktitle={ICCV},
  year={2023}
}

@article{siglip2,
  title={Siglip 2: Multilingual vision-language encoders with improved semantic understanding, localization, and dense features},
  author={Tschannen, Michael and Gritsenko, Alexey and Wang, Xiao and Naeem, Muhammad Ferjad and Alabdulmohsin, Ibrahim and Parthasarathy, Nikhil and Evans, Talfan and Beyer, Lucas and Xia, Ye and Mustafa, Basil and others},
  journal={arXiv preprint arXiv:2502.14786},
  year={2025}
}

@article{dinov2,
  title={Dinov2: Learning robust visual features without supervision},
  author={Oquab, Maxime and Darcet, Timoth{\'e}e and Moutakanni, Th{\'e}o and Vo, Huy and Szafraniec, Marc and Khalidov, Vasil and Fernandez, Pierre and Haziza, Daniel and Massa, Francisco and El-Nouby, Alaaeldin and others},
  journal={arXiv preprint arXiv:2304.07193},
  year={2023}
}

@inproceedings{k400,
  title={Quo vadis, action recognition? a new model and the kinetics dataset},
  author={Carreira, Joao and Zisserman, Andrew},
  booktitle={CVPR},
  year={2017}
}

@inproceedings{clip,
  title={Learning transferable visual models from natural language supervision},
  author={Radford, Alec and Kim, Jong Wook and Hallacy, Chris and Ramesh, Aditya and Goh, Gabriel and Agarwal, Sandhini and Sastry, Girish and Askell, Amanda and Mishkin, Pamela and Clark, Jack and others},
  booktitle={ICML},
  year={2021}
}

@inproceedings{dino,
  title={Emerging properties in self-supervised vision transformers},
  author={Caron, Mathilde and Touvron, Hugo and Misra, Ishan and J{\'e}gou, Herv{\'e} and Mairal, Julien and Bojanowski, Piotr and Joulin, Armand},
  booktitle={ICCV},
  year={2021}
}

@inproceedings{mae,
  title={Masked autoencoders are scalable vision learners},
  author={He, Kaiming and Chen, Xinlei and Xie, Saining and Li, Yanghao and Doll{\'a}r, Piotr and Girshick, Ross},
  booktitle={ICCV},
  year={2022}
}

@inproceedings{ijepa,
  title={Self-supervised learning from images with a joint-embedding predictive architecture},
  author={Assran, Mahmoud and Duval, Quentin and Misra, Ishan and Bojanowski, Piotr and Vincent, Pascal and Rabbat, Michael and LeCun, Yann and Ballas, Nicolas},
  booktitle={ICCV},
  year={2023}
}

@inproceedings{vivit,
  title={Vivit: A video vision transformer},
  author={Arnab, Anurag and Dehghani, Mostafa and Heigold, Georg and Sun, Chen and Lu{\v{c}}i{\'c}, Mario and Schmid, Cordelia},
  booktitle={ICCV},
  year={2021}
}

@inproceedings{timesformer,
  title={Is space-time attention all you need for video understanding?},
  author={Bertasius, Gedas and Wang, Heng and Torresani, Lorenzo},
  booktitle={ICML},
  year={2021}
}

@article{videomae,
  title={Videomae: Masked autoencoders are data-efficient learners for self-supervised video pre-training},
  author={Tong, Zhan and Song, Yibing and Wang, Jue and Wang, Limin},
  journal={NeurIPS},
  year={2022}
}

@inproceedings{dust3r,
  title={Dust3r: Geometric 3d vision made easy},
  author={Wang, Shuzhe and Leroy, Vincent and Cabon, Yohann and Chidlovskii, Boris and Revaud, Jerome},
  booktitle={CVPR},
  year={2024}
}

@inproceedings{fast3r,
  title={Fast3r: Towards 3d reconstruction of 1000+ images in one forward pass},
  author={Yang, Jianing and Sax, Alexander and Liang, Kevin J and Henaff, Mikael and Tang, Hao and Cao, Ang and Chai, Joyce and Meier, Franziska and Feiszli, Matt},
  booktitle={CVPR},
  year={2025}
}

@article{pi3,
  title={$\pi^{3}$: Permutation-Equivariant Visual Geometry Learning},
  author={Wang, Yifan and Zhou, Jianjun and Zhu, Haoyi and Chang, Wenzheng and Zhou, Yang and Li, Zizun and Chen, Junyi and Pang, Jiangmiao and Shen, Chunhua and He, Tong},
  journal={arXiv preprint arXiv:2507.13347},
  year={2025}
}

@article{d4rt,
  title={Efficiently Reconstructing Dynamic Scenes One D4RT at a Time},
  author={Zhang, Chuhan and Le Moing, Guillaume and Koppula, Skanda and Rocco, Ignacio and Momeni, Liliane and Xie, Junyu and Sun, Shuyang and Sukthankar, Rahul and Barral, Jo{\"e}lle K. and Hadsell, Raia and Ghahramani, Zoubin and Zisserman, Andrew and Zhang, Junlin and Sajjadi, Mehdi S. M.},
  journal={arXiv preprint},
  year={2025}
}

@inproceedings{span3r,
  title={3d reconstruction with spatial memory},
  author={Wang, Hengyi and Agapito, Lourdes},
  booktitle={3DV},
  year={2025},
}

@article{stream3r,
  title={Stream3r: Scalable sequential 3d reconstruction with causal transformer},
  author={Lan, Yushi and Luo, Yihang and Hong, Fangzhou and Zhou, Shangchen and Chen, Honghua and Lyu, Zhaoyang and Yang, Shuai and Dai, Bo and Loy, Chen Change and Pan, Xingang},
  journal={arXiv preprint arXiv:2508.10893},
  year={2025}
}

@article{point3r,
  title={Point3r: Streaming 3d reconstruction with explicit spatial pointer memory},
  author={Wu, Yuqi and Zheng, Wenzhao and Zhou, Jie and Lu, Jiwen},
  journal={arXiv preprint arXiv:2507.02863},
  year={2025}
}

@article{streamvggt,
  title={Streaming 4d visual geometry transformer},
  author={Zhuo, Dong and Zheng, Wenzhao and Guo, Jiahe and Wu, Yuqi and Zhou, Jie and Lu, Jiwen},
  journal={arXiv preprint arXiv:2507.11539},
  year={2025}
}

@inproceedings{sam,
  title={Segment anything},
  author={Kirillov, Alexander and Mintun, Eric and Ravi, Nikhila and Mao, Hanzi and Rolland, Chloe and Gustafson, Laura and Xiao, Tete and Whitehead, Spencer and Berg, Alexander C and Lo, Wan-Yen and others},
  booktitle={ICCV},
  year={2023}
}

@article{qwen3vl,
  title={Qwen3-VL Technical Report},
  author={Bai, Shuai and Cai, Yuxuan and Chen, Ruizhe and Chen, Keqin and Chen, Xionghui and Cheng, Zesen and Deng, Lianghao and Ding, Wei and Gao, Chang and Ge, Chunjiang and others},
  journal={arXiv preprint arXiv:2511.21631},
  year={2025}
}

@inproceedings{ibot,
  title={Image BERT Pre-training with Online Tokenizer},
  author={Zhou, Jinghao and Wei, Chen and Wang, Huiyu and Shen, Wei and Xie, Cihang and Yuille, Alan and Kong, Tao},
  booktitle={ICLR},
  year={2022}
}

@inproceedings{bonn,
  title={ReFusion: 3D reconstruction in dynamic environments for RGB-D cameras exploiting residuals},
  author={Palazzolo, Emanuele and Behley, Jens and Lottes, Philipp and Giguere, Philippe and Stachniss, Cyrill},
  booktitle={IROS},
  year={2019},
}

@inproceedings{tumd,
  title={A benchmark for the evaluation of RGB-D SLAM systems},
  author={Sturm, J{\"u}rgen and Engelhard, Nikolas and Endres, Felix and Burgard, Wolfram and Cremers, Daniel},
  booktitle={IROS},
  year={2012},
}

@inproceedings{sintel,
  title={A naturalistic open source movie for optical flow evaluation},
  author={Butler, Daniel J and Wulff, Jonas and Stanley, Garrett B and Black, Michael J},
  booktitle={ECCV},
  year={2012},
}

@article{kitti,
  title={Vision meets robotics: The kitti dataset},
  author={Geiger, Andreas and Lenz, Philip and Stiller, Christoph and Urtasun, Raquel},
  journal={International Journal of Robotics Research},
  year={2013},
  publisher={Sage Publications Sage UK: London}
}

@article{videommmu,
  title={Video-mmmu: Evaluating knowledge acquisition from multi-discipline professional videos},
  author={Hu, Kairui and Wu, Penghao and Pu, Fanyi and Xiao, Wang and Zhang, Yuanhan and Yue, Xiang and Li, Bo and Liu, Ziwei},
  journal={arXiv preprint arXiv:2501.13826},
  year={2025}
}

@article{perceptiontest,
  title={Perception test: A diagnostic benchmark for multimodal video models},
  author={Patraucean, Viorica and Smaira, Lucas and Gupta, Ankush and Recasens, Adria and Markeeva, Larisa and Banarse, Dylan and Koppula, Skanda and Malinowski, Mateusz and Yang, Yi and Doersch, Carl and others},
  journal={NeurIPS},
  year={2023}
}

@article{egoschema,
  title={Egoschema: A diagnostic benchmark for very long-form video language understanding},
  author={Mangalam, Karttikeya and Akshulakov, Raiymbek and Malik, Jitendra},
  journal={NeurIPS},
  year={2023}
}

@article{koleo,
  title={Spreading vectors for similarity search},
  author={Sablayrolles, Alexandre and Douze, Matthijs and Schmid, Cordelia and J{\'e}gou, Herv{\'e}},
  journal={arXiv preprint arXiv:1806.03198},
  year={2018}
}

@article{gpt4o,
  title={Gpt-4o system card},
  author={Hurst, Aaron and Lerer, Adam and Goucher, Adam P and Perelman, Adam and Ramesh, Aditya and Clark, Aidan and Ostrow, AJ and Welihinda, Akila and Hayes, Alan and Radford, Alec and others},
  journal={arXiv preprint arXiv:2410.21276},
  year={2024}
}

@article{gemini_1.5,
  title={Gemini 1.5: Unlocking multimodal understanding across millions of tokens of context},
  author={Team, Gemini and Georgiev, Petko and Lei, Ving Ian and Burnell, Ryan and Bai, Libin and Gulati, Anmol and Tanzer, Garrett and Vincent, Damien and Pan, Zhufeng and Wang, Shibo and others},
  journal={arXiv preprint arXiv:2403.05530},
  year={2024}
}

@article{llava_one_vision,
  title={Llava-onevision: Easy visual task transfer},
  author={Li, Bo and Zhang, Yuanhan and Guo, Dong and Zhang, Renrui and Li, Feng and Zhang, Hao and Zhang, Kaichen and Zhang, Peiyuan and Li, Yanwei and Liu, Ziwei and others},
  journal={arXiv preprint arXiv:2408.03326},
  year={2024}
}

@article{llavavideo,
  title={Llava-video: Video instruction tuning with synthetic data},
  author={Zhang, Yuanhan and Wu, Jinming and Li, Wei and Li, Bo and Ma, Zejun and Liu, Ziwei and Li, Chunyuan},
  journal={arXiv preprint arXiv:2410.02713},
  year={2024}
}

@article{qwen25vl,
  title={Qwen2. 5-vl technical report},
  author={Bai, Shuai and Chen, Keqin and Liu, Xuejing and Wang, Jialin and Ge, Wenbin and Song, Sibo and Dang, Kai and Wang, Peng and Wang, Shijie and Tang, Jun and others},
  journal={arXiv preprint arXiv:2502.13923},
  year={2025}
}

@article{spacer,
  title={Spacer: Reinforcing mllms in video spatial reasoning},
  author={Ouyang, Kun and Liu, Yuanxin and Wu, Haoning and Liu, Yi and Zhou, Hao and Zhou, Jie and Meng, Fandong and Sun, Xu},
  journal={arXiv preprint arXiv:2504.01805},
  year={2025}
}

@article{VILASR,
  title={Reinforcing spatial reasoning in vision-language models with interwoven thinking and visual drawing},
  author={Wu, Junfei and Guan, Jian and Feng, Kaituo and Liu, Qiang and Wu, Shu and Wang, Liang and Wu, Wei and Tan, Tieniu},
  journal={arXiv preprint arXiv:2506.09965},
  year={2025}
}

@article{vlm3r,
  title={Vlm-3r: Vision-language models augmented with instruction-aligned 3d reconstruction},
  author={Fan, Zhiwen and Zhang, Jian and Li, Renjie and Zhang, Junge and Chen, Runjin and Hu, Hezhen and Wang, Kevin and Qu, Huaizhi and Wang, Dilin and Yan, Zhicheng and others},
  journal={arXiv preprint arXiv:2505.20279},
  year={2025}
}

@article{vst,
  title={Visual spatial tuning},
  author={Yang, Rui and Zhu, Ziyu and Li, Yanwei and Huang, Jingjia and Yan, Shen and Zhou, Siyuan and Liu, Zhe and Li, Xiangtai and Li, Shuangye and Wang, Wenqian and others},
  journal={arXiv preprint arXiv:2511.05491},
  year={2025}
}

@article{spacemind,
  title={SpaceMind: Camera-Guided Modality Fusion for Spatial Reasoning in Vision-Language Models},
  author={Zhao, Ruosen and Zhang, Zhikang and Xu, Jialei and Chang, Jiahao and Chen, Dong and Li, Lingyun and Sun, Weijian and Wei, Zizhuang},
  journal={arXiv preprint arXiv:2511.23075},
  year={2025}
}

@misc{qwen25,
      title={Qwen2.5 Technical Report}, 
      author={Qwen and : and An Yang and Baosong Yang and Beichen Zhang and Binyuan Hui and Bo Zheng and Bowen Yu and Chengyuan Li and Dayiheng Liu and Fei Huang and Haoran Wei and Huan Lin and Jian Yang and Jianhong Tu and Jianwei Zhang and Jianxin Yang and Jiaxi Yang and Jingren Zhou and Junyang Lin and Kai Dang and Keming Lu and Keqin Bao and Kexin Yang and Le Yu and Mei Li and Mingfeng Xue and Pei Zhang and Qin Zhu and Rui Men and Runji Lin and Tianhao Li and Tianyi Tang and Tingyu Xia and Xingzhang Ren and Xuancheng Ren and Yang Fan and Yang Su and Yichang Zhang and Yu Wan and Yuqiong Liu and Zeyu Cui and Zhenru Zhang and Zihan Qiu},
      year={2025},
      eprint={2412.15115},
      archivePrefix={arXiv},
      primaryClass={cs.CL},
      url={https://arxiv.org/abs/2412.15115}, 
}

@article{vlm4vla,
  title={VLM4VLA: Revisiting Vision-Language-Models in Vision-Language-Action Models},
  author={Zhang, Jianke and Chen, Xiaoyu and Wang, Qiuyue and Li, Mingsheng and Guo, Yanjiang and Hu, Yucheng and Zhang, Jiajun and Bai, Shuai and Lin, Junyang and Chen, Jianyu},
  journal={arXiv preprint arXiv:2601.03309},
  year={2026}
}

@article{robovlm,
  title={Towards Generalist Robot Policies: What Matters in Building Vision-Language-Action Models},
  author={Li, Xinghang and Li, Peiyan and Liu, Minghuan and Wang, Dong and Liu, Jirong and Kang, Bingyi and Ma, Xiao and Kong, Tao and Zhang, Hanbo and Liu, Huaping},
  journal={arXiv preprint arXiv:2412.14058},
  year={2024}
}

@article{florence,
  title={Florence: A new foundation model for computer vision},
  author={Yuan, Lu and Chen, Dongdong and Chen, Yi-Ling and Codella, Noel and Dai, Xiyang and Gao, Jianfeng and Hu, Houdong and Huang, Xuedong and Li, Boxin and Li, Chunyuan and others},
  journal={arXiv preprint arXiv:2111.11432},
  year={2021}
}

@inproceedings{florence2,
  title={Florence-2: Advancing a unified representation for a variety of vision tasks},
  author={Xiao, Bin and Wu, Haiping and Xu, Weijian and Dai, Xiyang and Hu, Houdong and Lu, Yumao and Zeng, Michael and Liu, Ce and Yuan, Lu},
  booktitle={CVPR},
  year={2024}
}

@inproceedings{ofa,
  title={Ofa: Unifying architectures, tasks, and modalities through a simple sequence-to-sequence learning framework},
  author={Wang, Peng and Yang, An and Men, Rui and Lin, Junyang and Bai, Shuai and Li, Zhikang and Ma, Jianxin and Zhou, Chang and Zhou, Jingren and Yang, Hongxia},
  booktitle={ICML},
  year={2022},
}

@article{unified-io,
  title={Unified-io: A unified model for vision, language, and multi-modal tasks},
  author={Lu, Jiasen and Clark, Christopher and Zellers, Rowan and Mottaghi, Roozbeh and Kembhavi, Aniruddha},
  journal={arXiv preprint arXiv:2206.08916},
  year={2022}
}

@inproceedings{unified-io2,
  title={Unified-io 2: Scaling autoregressive multimodal models with vision language audio and action},
  author={Lu, Jiasen and Clark, Christopher and Lee, Sangho and Zhang, Zichen and Khosla, Savya and Marten, Ryan and Hoiem, Derek and Kembhavi, Aniruddha},
  booktitle={CVPR},
  year={2024}
}

@inproceedings{adam,
  author    = {Kingma, Diederik P. and Ba, Jimmy},
  title     = {Adam: A Method for Stochastic Optimization},
  booktitle = {ICLR},
  year      = {2015},
}

@article{qwenvl,
  title={Qwen technical report},
  author={Bai, Jinze and Bai, Shuai and Chu, Yunfei and Cui, Zeyu and Dang, Kai and Deng, Xiaodong and Fan, Yang and Ge, Wenbin and Han, Yu and Huang, Fei and others},
  journal={arXiv preprint arXiv:2309.16609},
  year={2023}
}

@article{llava,
  title={Visual instruction tuning},
  author={Liu, Haotian and Li, Chunyuan and Wu, Qingyang and Lee, Yong Jae},
  journal={NeurIPS},
  year={2023}
}

@misc{llavanext,
    title={LLaVA-NeXT: Improved reasoning, OCR, and world knowledge},
    url={https://llava-vl.github.io/blog/2024-01-30-llava-next/},
    author={Liu, Haotian and Li, Chunyuan and Li, Yuheng and Li, Bo and Zhang, Yuanhan and Shen, Sheng and Lee, Yong Jae},
    month={January},
    year={2024}
}

@inproceedings{clipblind,
  title={Eyes wide shut? exploring the visual shortcomings of multimodal llms},
  author={Tong, Shengbang and Liu, Zhuang and Zhai, Yuexiang and Ma, Yi and LeCun, Yann and Xie, Saining},
  booktitle={CVPR},
  year={2024}
}

@article{llava-ov-1.5,
  title={Llava-onevision-1.5: Fully open framework for democratized multimodal training},
  author={An, Xiang and Xie, Yin and Yang, Kaicheng and Zhang, Wenkang and Zhao, Xiuwei and Cheng, Zheng and Wang, Yirui and Xu, Songcen and Chen, Changrui and Zhu, Didi and others},
  journal={arXiv preprint arXiv:2509.23661},
  year={2025}
}

@inproceedings{cambrian,
  title={Cambrian-1: A fully open, vision-centric exploration of multimodal llms},
  author={Tong, Peter and Brown, Ellis and Wu, Penghao and Woo, Sanghyun and IYER, Adithya Jairam Vedagiri and Akula, Sai Charitha and Yang, Shusheng and Yang, Jihan and Middepogu, Manoj and Wang, Ziteng and others},
  booktitle={NeurIPS},
  year={2024}
}

@article{deepseek_vl,
  title={Deepseek-vl: towards real-world vision-language understanding},
  author={Lu, Haoyu and Liu, Wen and Zhang, Bo and Wang, Bingxuan and Dong, Kai and Liu, Bo and Sun, Jingxiang and Ren, Tongzheng and Li, Zhuoshu and Yang, Hao and others},
  journal={arXiv preprint arXiv:2403.05525},
  year={2024}
}

@inproceedings{brave,
  title={Brave: Broadening the visual encoding of vision-language models},
  author={Kar, O{\u{g}}uzhan Fatih and Tonioni, Alessio and Poklukar, Petra and Kulshrestha, Achin and Zamir, Amir and Tombari, Federico},
  booktitle={ECCV},
  year={2024},
}

@article{eagle,
  title={Eagle: Exploring the design space for multimodal llms with mixture of encoders},
  author={Shi, Min and Liu, Fuxiao and Wang, Shihao and Liao, Shijia and Radhakrishnan, Subhashree and Zhao, Yilin and Huang, De-An and Yin, Hongxu and Sapra, Karan and Yacoob, Yaser and others},
  journal={arXiv preprint arXiv:2408.15998},
  year={2024}
}

@inproceedings{rt2,
  title={Rt-2: Vision-language-action models transfer web knowledge to robotic control},
  author={Zitkovich, Brianna and Yu, Tianhe and Xu, Sichun and Xu, Peng and Xiao, Ted and Xia, Fei and Wu, Jialin and Wohlhart, Paul and Welker, Stefan and Wahid, Ayzaan and others},
  booktitle={CoRL},
  year={2023},
}

@article{pi0,
  title={$\pi_0 $: A Vision-Language-Action Flow Model for General Robot Control},
  author={Black, Kevin and Brown, Noah and Driess, Danny and Esmail, Adnan and Equi, Michael and Finn, Chelsea and Fusai, Niccolo and Groom, Lachy and Hausman, Karol and Ichter, Brian and others},
  journal={arXiv preprint arXiv:2410.24164},
  year={2024}
}

@article{openvla,
    title={OpenVLA: An Open-Source Vision-Language-Action Model},
    author={{Moo Jin} Kim and Karl Pertsch and Siddharth Karamcheti and Ted Xiao and Ashwin Balakrishna and Suraj Nair and Rafael Rafailov and Ethan Foster and Grace Lam and Pannag Sanketi and Quan Vuong and Thomas Kollar and Benjamin Burchfiel and Russ Tedrake and Dorsa Sadigh and Sergey Levine and Percy Liang and Chelsea Finn},
    journal = {arXiv preprint arXiv:2406.09246},
    year={2024},
}

@article{paligemma,
  title={Paligemma: A versatile 3b vlm for transfer},
  author={Beyer, Lucas and Steiner, Andreas and Pinto, Andr{\'e} Susano and Kolesnikov, Alexander and Wang, Xiao and Salz, Daniel and Neumann, Maxim and Alabdulmohsin, Ibrahim and Tschannen, Michael and Bugliarello, Emanuele and others},
  journal={arXiv preprint arXiv:2407.07726},
  year={2024}
}

@article{paligemma2,
  title={Paligemma 2: A family of versatile vlms for transfer},
  author={Steiner, Andreas and Pinto, Andr{\'e} Susano and Tschannen, Michael and Keysers, Daniel and Wang, Xiao and Bitton, Yonatan and Gritsenko, Alexey and Minderer, Matthias and Sherbondy, Anthony and Long, Shangbang and others},
  journal={arXiv preprint arXiv:2412.03555},
  year={2024}
}

@article{qwen3,
  title={Qwen3 technical report},
  author={Yang, An and Li, Anfeng and Yang, Baosong and Zhang, Beichen and Hui, Binyuan and Zheng, Bo and Yu, Bowen and Gao, Chang and Huang, Chengen and Lv, Chenxu and others},
  journal={arXiv preprint arXiv:2505.09388},
  year={2025}
}

@article{thinkact,
  title={Thinkact: Vision-language-action reasoning via reinforced visual latent planning},
  author={Huang, Chi-Pin and Wu, Yueh-Hua and Chen, Min-Hung and Wang, Yu-Chiang Frank and Yang, Fu-En},
  journal={arXiv preprint arXiv:2507.16815},
  year={2025}
}

@InProceedings{streamformer,
    author    = {Yan, Yibin and Xu, Jilan and Di, Shangzhe and Liu, Yikun and Shi, Yudi and Chen, Qirui and Li, Zeqian and Huang, Yifei and Xie, Weidi},
    title     = {Learning Streaming Video Representation via Multitask Training},
    booktitle = {ICCV},
    year      = {2025},
}

@inproceedings{super-instruction,
  title={Super-naturalinstructions: Generalization via declarative instructions on 1600+ nlp tasks},
  author={Wang, Yizhong and Mishra, Swaroop and Alipoormolabashi, Pegah and Kordi, Yeganeh and Mirzaei, Amirreza and Naik, Atharva and Ashok, Arjun and Dhanasekaran, Arut Selvan and Arunkumar, Anjana and Stap, David and others},
  booktitle={EMNLP},
  year={2022}
}

@article{gpt3,
  title={Language models are few-shot learners},
  author={Brown, Tom and Mann, Benjamin and Ryder, Nick and Subbiah, Melanie and Kaplan, Jared D and Dhariwal, Prafulla and Neelakantan, Arvind and Shyam, Pranav and Sastry, Girish and Askell, Amanda and others},
  journal={NeurIPS},
  year={2020}
}

@article{gpt4,
  title={Gpt-4 technical report},
  author={Achiam, Josh and Adler, Steven and Agarwal, Sandhini and Ahmad, Lama and Akkaya, Ilge and Aleman, Florencia Leoni and Almeida, Diogo and Altenschmidt, Janko and Altman, Sam and Anadkat, Shyamal and others},
  journal={arXiv preprint arXiv:2303.08774},
  year={2023}
}

@inproceedings{imagenet,
  title={Imagenet: A large-scale hierarchical image database},
  author={Deng, Jia and Dong, Wei and Socher, Richard and Li, Li-Jia and Li, Kai and Fei-Fei, Li},
  booktitle={CVPR},
  year={2009},
}

@inproceedings{vsibench,
  title={Thinking in space: How multimodal large language models see, remember, and recall spaces},
  author={Yang, Jihan and Yang, Shusheng and Gupta, Anjali W and Han, Rilyn and Fei-Fei, Li and Xie, Saining},
  booktitle={CVPR},
  year={2025}
}

@article{calvin,
  title={Calvin: A benchmark for language-conditioned policy learning for long-horizon robot manipulation tasks},
  author={Mees, Oier and Hermann, Lukas and Rosete-Beas, Erick and Burgard, Wolfram},
  journal={IEEE-RAL},
  year={2022},
  publisher={IEEE}
}

@inproceedings{simplerenv,
  title={Evaluating Real-World Robot Manipulation Policies in Simulation},
  author={Li, Xuanlin and Hsu, Kyle and Gu, Jiayuan and Mees, Oier and Pertsch, Karl and Walke, Homer Rich and Fu, Chuyuan and Lunawat, Ishikaa and Sieh, Isabel and Kirmani, Sean and others},
  booktitle={CoRL},
  year={2024} 
}

@inproceedings{videomme,
  title={Video-mme: The first-ever comprehensive evaluation benchmark of multi-modal llms in video analysis},
  author={Fu, Chaoyou and Dai, Yuhan and Luo, Yongdong and Li, Lei and Ren, Shuhuai and Zhang, Renrui and Wang, Zihan and Zhou, Chenyu and Shen, Yunhang and Zhang, Mengdan and others},
  booktitle={CVPR},
  year={2025}
}

@inproceedings{ade20k,
  title={Scene parsing through ade20k dataset},
  author={Zhou, Bolei and Zhao, Hang and Puig, Xavier and Fidler, Sanja and Barriuso, Adela and Torralba, Antonio},
  booktitle={CVPR},
  year={2017}
}

@inproceedings{NYUv2,
  title={Indoor segmentation and support inference from rgbd images},
  author={Silberman, Nathan and Hoiem, Derek and Kohli, Pushmeet and Fergus, Rob},
  booktitle={ECCV},
  year={2012},
}

@article{davis17,
  title={The 2017 davis challenge on video object segmentation},
  author={Pont-Tuset, Jordi and Perazzi, Federico and Caelles, Sergi and Arbel{\'a}ez, Pablo and Sorkine-Hornung, Alex and Van Gool, Luc},
  journal={arXiv preprint arXiv:1704.00675},
  year={2017}
}

@article{cambrian-s,
  title={Cambrian-s: Towards spatial supersensing in video},
  author={Yang, Shusheng and Yang, Jihan and Huang, Pinzhi and Brown, Ellis and Yang, Zihao and Yu, Yue and Tong, Shengbang and Zheng, Zihan and Xu, Yifan and Wang, Muhan and others},
  journal={arXiv preprint arXiv:2511.04670},
  year={2025}
}

@article{vica,
  title={Towards visuospatial cognition via hierarchical fusion of visual experts},
  author={Feng, Qi},
  journal={arXiv preprint arXiv:2505.12363},
  year={2025}
}

@inproceedings{anetqa,
    author = {Yu, Zhou and Xu, Dejing and Yu, Jun and Yu, Ting and Zhao, Zhou and Zhuang, Yueting and Tao, Dacheng},
    title = {ActivityNet-QA: A Dataset for Understanding Complex Web Videos via Question Answering},
    booktitle = {AAAI},
    year = {2019}
}

@InProceedings{nextqa,
    author    = {Xiao, Junbin and Shang, Xindi and Yao, Angela and Chua, Tat-Seng},
    title     = {NExT-QA: Next Phase of Question-Answering to Explaining Temporal Actions},
    booktitle = {CVPR},
    year      = {2021},
}

@article{llavahound,
  title={Direct Preference Optimization of Video Large Multimodal Models from Language Model Reward},
  author={Zhang, Ruohong and Gui, Liangke and Sun, Zhiqing and Feng, Yihao and Xu, Keyang and Zhang, Yuanhan and Fu, Di and Li, Chunyuan and Hauptmann, Alexander and Bisk, Yonatan and others},
  journal={arXiv preprint arXiv:2404.01258},
  year={2024}
}

@article{flashattn,
  title={Flashattention: Fast and memory-efficient exact attention with io-awareness},
  author={Dao, Tri and Fu, Dan and Ermon, Stefano and Rudra, Atri and R{\'e}, Christopher},
  journal={NeurIPS},
  year={2022}
}
}

\end{document}